# Physics of Language Models: Part 3.3, Knowledge Capacity Scaling Laws

Zeyuan Allen-Zhu
zeyuanallenzhu@meta.com
Meta / FAIR Labs

Yuanzhi Li
Yuanzhi.Li@mbzuai.ac.ae
Mohamed bin Zayed University of AI

April 7, 2024

(version 1)*

## Abstract

Scaling laws describe the relationship between the size of language models and their capabilities. Unlike prior studies that evaluate a model's capability via loss or benchmarks, we estimate the number of knowledge *bits* a model stores. We focus on factual knowledge represented as tuples, such as (USA, capital, Washington D.C.) from a Wikipedia page. Through multiple controlled datasets, we establish that language models can and only can store **2 bits of knowledge per parameter, even when quantized to int8**, and such knowledge can be flexibly extracted for downstream applications. Consequently, a 7B model can store 14B bits of knowledge, surpassing the English Wikipedia and textbooks combined based on our estimation.

**More broadly, we present 12 results** on how (1) training duration, (2) model architecture, (3) quantization, (4) sparsity constraints such as MoE, and (5) data signal-to-noise ratio affect a model's knowledge storage capacity. Notable insights include:

- The GPT-2 architecture, with rotary embedding, matches or even surpasses LLaMA/Mistral architectures *in knowledge storage*, particularly over shorter training durations. This arises because LLaMA/Mistral uses GatedMLP, which is less stable and harder to train.

- Prepending training data with domain names (e.g., wikipedia.org) significantly increases a model's knowledge capacity. Language models can autonomously identify and prioritize domains rich in knowledge, optimizing their storage capacity.

---

*Submitted for Meta internal review on March 14, 2024. We would like to thank Lin Xiao and Yuchen Zhang for many helpful conversations. We would like to extend special thanks to Ian Clark, Gourab De, Anmol Mann, and Max Pfeifer from W&B, as well as Lucca Bertoncini, Liao Hu, Caleb Ho, Apostolos Kokolis, and Shubho Sengupta from Meta FAIR NextSys; Henry Estela, Wil Johnson, Rizwan Hashmi, and Lucas Noah from Meta Cloud Foundation; without their invaluable support, the extensive experiments in this paper would not have been possible.

# 1  Introduction

The scaling laws of large language models remain a pivotal area of research, enabling predictions about the performance of extremely large models through experiments with smaller ones. On the training time aspect, established scaling laws [1, 13, 14, 16, 21] discuss the optimal training flops versus model size. However, recent studies [12, 24, 25] challenge these laws, demonstrating that training smaller models with significantly more flops can yield superior results. While these laws talk about how much time/data is needed to train a model of a certain size, another fundamental question is: *what is the ultimate performance a model can achieve, assuming sufficient training?* Despite the known emergent behaviors in large models [8, 34], there is a *lack of a principled, quantitative analysis* on how model size impacts its capacity when adequately trained.[1]

Traditional theory on overparameterization suggests that scaling up model size in sufficiently trained models can enhance memorization of training data [6], improve generalization error [15, 27, 28], and better fit complex target functions [5, 23]. However, these results often overlook large constant or polynomial factors, leading to a significant discrepancy from practical outcomes.

In this paper, we introduce a principled framework to examine *highly accurate* scaling laws concerning model size versus its *knowledge storage capacity*. It is intuitive that larger language models can store more knowledge, but does the total knowledge scale linearly with the model's size? What is the **exact constant** of this scaling? Understanding this constant is crucial for assessing the efficiency of transformer models in knowledge storage and how various factors (e.g., architecture, quantization, training duration, etc.) influence this capacity.

Knowledge is a, if not the, pivotal component of human intelligence, accumulated over our extensive history. Large language models like GPT-4 are celebrated not just for their sophisticated logic but also for their superior knowledge base. Despite rumors of GPT-4 having over 1T parameters, *is it necessary to store all human knowledge?* Could a 10B model, if trained sufficiently with high-quality data, match GPT-4's knowledge capacity? Our paper seeks to address these questions.

**Knowledge Pieces.** Defining "one piece of human knowledge" precisely is challenging. This paper aims to make progress by focusing on a restricted, yet sufficiently interesting domain. We define a *piece* of knowledge as a (name, attribute, value) tuple, e.g., (Anya Forger, birthday, 10/2/1996); and many data in world knowledge benchmarks can be broken down into pieces like this.[2]

We generate *synthetic* knowledge-only datasets by uniformly at random generating (name, attribute, value) tuples from a knowledge base and converting them into English descriptions. We pretrain language models (e.g., GPT-2, LLaMA, Mistral) on these texts using a standard autoregressive objective from random initialization, and "estimate" the learned knowledge. By varying the number of knowledge pieces and model sizes, we outline a knowledge capacity scaling law.

Our idealized setting, free from irrelevant data, allows for more accurate scaling law computations — we also discuss how "junk" data affects capacity later in Section 10. In contrast, it is difficult to quantify real-life knowledge; for instance, if LLaMA-70B outperforms LLaMA-7B by 30% on a benchmark, it doesn't necessarily mean a tenfold model scaling only boosts capacity by 30% (see Footnote 1). The synthetic setting also lets us adjust various hyperparameters, like

---

[1]There is a rich literature comparing how pretrained models perform on benchmark tasks. Most comparisons are for different model families trained over different data: if LLaMA-70B is better than Mistral-7B, does the gain come from its choice of pretrain data, or the architecture difference, or really the size of the model? Some comparisons are among the same architecture, such as LLaMA-70B scores 63.6% on the world knowledge benchmark while LLaMA-7B scores only 48.9% [33]; does this mean increasing model size by 10x increases its capacity only to $130\% = 63.6/48.9$? Thus, it is highly important to use a more principled framework to study scaling laws in a controlled setting.

[2]Examples include (Africa, largest country, Sudan) and (It Happened One Night, director, Frank Capra) in TriviaQA [20], or (Teton Dam, collapse date, 06/05/1976) and (USA, Capital, Washington D.C.) in NaturalQuestions [22].

name/value lengths and vocabulary size, to study their effects on knowledge capacity scaling laws.

Most of the paper shall focus on a setting with synthetically-generated human biographies as data, either using predefined sentence templates or LLaMA2-generated biographies for realism.

**Bit Complexity and Capacity Ratio.** For $N$ knowledge pieces (i.e., $N$ tuples), we define the *bit complexity* as the minimum bits required to encode these tuples. For any language model trained on this data, we calculate its "bit complexity lower bound" (see Theorem 3.2), describing the minimum number of bits needed for the model to store the knowledge at its given accuracy. This formula is nearly as precise as the upper bound, within a $1 - o(1)$ factor.

We train language models of varying sizes on knowledge data with different $N$ values. By comparing the models' trainable parameters to the bit complexity lower bounds, we evaluate their knowledge storage efficiency.

*Example.* A model with 100M parameters storing 220M bits of knowledge has a *capacity ratio* of $2.2 = \frac{220\text{M}}{100\text{M}}$ bits per parameter. There is also a trivial upper bound on a model's capacity ratio; for instance, a model using int8 parameters cannot exceed a capacity ratio of 8.

**Our results.** Our findings are summarized as follows:

- SECTION 5: BASE SCALING LAW FOR GPT2. [3]

  - RESULT 1+2+3 : GPT2, trained with standard AdamW, consistently achieves a 2bit/param capacity ratio across all data settings after sufficient training. This includes various model sizes, depths, widths, data sizes, types (synthetic/semi-synthetic), and hyperparameters (e.g., name/value length, attribute number, value diversity).

  *Remark* 1.1. This predicts **a sufficiently trained 7B language model** can store 14B bits of knowledge, surpassing the knowledge of English Wikipedia and textbooks by our estimation.[4]

  *Remark* 1.2. When we say the model *stores knowledge*, it isn't word-by-word memorization. Instead, the knowledge is flexibly extractable (e.g., via QAs like "What is Anya Forger's birthday") [3] and applicable in downstream tasks (e.g., comparing birthdays) via fine-tune [4].

- SECTION 6: HOW TRAINING TIME AFFECTS MODEL CAPACITY.

  Achieving a 2bit/param capacity requires each knowledge piece to be visited 1000 times during training, termed **1000-exposure** to differentiate from traditional "1000-pass" terminology, as a single data pass can expose a knowledge piece 1000 times.[5]

  - RESULT 4 : With 100 exposures, an *undertrained* GPT2's capacity ratio falls to 1bit/param.

  *Remark* 1.3. Another perspective on Result 4 is that *rare* knowledge, encountered only 100 times during training, is stored at a 1bit/param ratio.

- SECTION 7: HOW MODEL ARCHITECTURE AFFECTS MODEL CAPACITY.

  We tested LLaMA, Mistral, and GPT2 architectures with reduced or even no MLP layers.

---

[3]In this paper, GPT2 refers to the original GPT2 model but with rotary embedding instead of positional embedding and without dropout.

[4]As of February 1, 2024, English Wikipedia contains a total of 4.5 billion words, see https://en.wikipedia.org/wiki/Wikipedia:Size_of_Wikipedia#Size_of_the_English_Wikipedia_database, accessed March 2024. We estimate that the non-overlapping contents of English textbooks have fewer than 16 billion words in total, see Remark G.1. This amounts to 20.5 billion words, and we believe they contain fewer than 14 billion bits of knowledge.

[5]For example, it is plausible that one pass through Wiki data might present the knowledge piece (US, capital, Washington D.C.) 1000 times, and one pass through the Common Crawl might present it a million times.

- RESULT 5: In the 1000-exposure setting, a 2bit/param capacity ratio appears to be a **universal rule**: all models, even without MLP layers, closely achieve this ratio.
- RESULT 6: With 100 exposures, some archs show limitations; notably, LLaMA/Mistral's capacity ratio is 1.3x lower than GPT2's, even after best-tuned learning rates.
- RESULT 7: Further controlled experiments indicate that "gated MLP" usage leads to LLaMA/Mistral architecture's underperformance in knowledge storage.

*Remark* 1.4. **Our framework offers a principled playground to compare models.** This contrasts with traditional comparisons based on loss/perplexity, which can produce debatable conclusions.[6] Controlled data also reveal more significant differences between models.[7]

- SECTION 8: HOW QUANTIZATION AFFECTS MODEL CAPACITY.

  We applied `GPTQ` [10] to quantize models from the base scaling laws to int8 or int4. Surprisingly,

  - RESULT 8: Quantizing to int8 does not compromise model capacity (even for models on the boundary of 2bit/param); however, quantizing to int4 reduces capacity to 0.7bit/param.

  *Remark* 1.5. Since int8 is 8bit, LLMs can exceed 1/4 of the theoretical limit for storing knowledge; thus knowledge must be very compactly stored inside the model across all layers.

  *Remark* 1.6. Since 2bit/param is obtained after sufficient training, training longer *may not* further improve model capacity, *but quantization can.* While not covered in this paper, our framework also provides a principled playground to compare different quantization methods.

- SECTION 9: HOW SPARSITY (MoE) AFFECTS MODEL CAPACITY.

  Mixture-of-experts (MoE) models offer faster inference than dense models but often underperform dense models with the same total parameter count (not effective parameters). We show that this performance drop is likely not due to a lack of knowledge storage capability.

  - RESULT 9: MoE models, even with 32 experts, only reduce 1.3x in capacity compared to the base scaling laws, despite using just 8.8% of the total parameters during inference.

- Section 10: HOW JUNK KNOWLEDGE AFFECTS MODEL CAPACITY.

  Not all pretrain data are equally useful. Much of the internet data lacks valuable knowledge for training language models [24], while knowledge-rich sources like Wikipedia represent only a small fraction of the training tokens. We explore the impact on model capacity by conducting a controlled experiment with both useful and "junk" data.

  - RESULT 10+11: Junk data significantly reduces model capacity. As an example, with a 1:7 ratio of "useful to junk" training tokens, capacity for useful knowledge *loses by a factor of 20x*, even when useful knowledge is exposed 100 times.[8]

---

[6] A model might achieve better perplexity by performing *much better* on simpler data but slightly poorer on complex data, or by excelling in reasoning tasks but not in knowledge storage. Our results offer a more nuanced view: GatedMLP doesn't affect frequently encountered knowledge (with 1000 exposures) but does impact moderately rare knowledge (with 100 exposures).

[7] For example, Shazeer [29] found GatedMLP offers a $\sim 1\%$ accuracy boost on benchmark tasks; our findings of a 1.3x difference translates for instance to accuracies 90% vs. 70%.

[8] The loss factor improves to 3x/1.5x/1.3x with 300/600/1000 exposures of useful knowledge, compared to Result 4 which involves training without junk for only 100 exposures.

- RESULT 12 : An *effective mitigation* is to prepend a special token to all useful knowledge. This is akin to adding a domain name like wikipedia.org at the start of every Wikipedia paragraph; the model **autonomously** identifies high-quality data without prior knowledge of valuable domains. In the example above, the loss factor improves from 20x to 2x.

Overall, our approach to studying knowledge capacity scaling laws offers a flexible and **more accurate playground** compared to traditional methods that evaluate language models trained on internet data against real-world benchmarks. This accuracy is partly due to the synthetic nature of our dataset, which eliminates concerns about benchmark contamination that could compromise the validity of real-world benchmark results. In this paper, we've conducted a thorough comparison across different model architectures and types of knowledge. While we haven't explored various quantization methods, this represents a promising direction for future research. We've also investigated the impact of junk data and proposed mitigation strategies. We believe the insights gained from this principled exploration can assist practitioners in making informed decisions about model selection, training data preparation, and further theoretical research into LLMs.

## 2 Preliminaries

In this paper, a piece of knowledge is a tuple of three strings: (name, attribute, value) = $(n, a, v)$. For instance, $n =$ "Anya", $a =$ "birthday", $v =$ "Oct 2, 1996".

### 2.1 Knowledge (Theoretical Setting)

The complexity of a knowledge set is determined not only by the number of knowledge pieces but also by the length of the value string $v$, the diversity of the vocabulary, and other factors. For instance, if the attribute $a =$ "passport number," then the value $v$ contains more bits of knowledge compared with $a =$ "gender," because the former has significantly higher *diversity*. If the attribute $a =$ "birth date," then the value $v$ could consist of 3 *chunks*: $(10, 2, 1996)$.

Considering these examples, we propose a set of hyperparameters that may influence the complexity of knowledge:

1. $N$ — the number of (distinct) names $n$, denoted by $\mathcal{N}$.

2. $K$ — the number of attributes $a$, with $\mathcal{A}$ representing the set of attributes. For simplicity, we assume $|\mathcal{A}| = K$ is fixed.

3. $T$ — the number of tokens $T$, where every character in $v$ belongs to $\mathcal{T}$ for some $|\mathcal{T}| = T$. For example, we can think of $T$ as "vocab size" in a tokenizer.

4. $C$ and $L$ — the number of chunks and the length of each chunk for the value: each value $v \in (\mathcal{T}^L)^C$ can be expressed as $v = (v_1, v_2, \cdots, v_C)$, where $v_i \in \mathcal{T}^L$.

5. $D$ — the diversity of chunks: for each piece of knowledge $(n, a, v)$ and $i \in [C]$, the chunk $v_i$ belongs to $\mathcal{D}_a \subset \mathcal{T}^L$, for some set with cardinality $D \stackrel{\text{def}}{=} |\mathcal{D}_a| \ll T^L$.

*Remark* 2.1. For notation simplicity, we have assumed that all chunks within an attribute $a \in \mathcal{A}$ share the same diversity set $\mathcal{D}_a$, and all chunks are of equal length, etc. This enables us to more easily demonstrate the influence of each hyperparameter on a model's capacity. In practice, different attributes may have different diversity sets or value lengths — e.g., $\mathcal{D}_{\text{passport}}$ could be much larger than $\mathcal{D}_{\text{gender}}$. Our theoretical results do apply to these settings, albeit with more complex notation.

In our theoretical result, we introduce a dataset $\mathsf{bioD}(N, K, C, D, L, T)$ defined as follows:

**Definition 2.2** (bioD data generation). *Consider a fixed set of $K$ attributes, such as a set $\mathcal{A} = \{$"ID 1" ... "ID K"$\}$, and a fixed set $\mathcal{N}_0$ of candidate names (with $N_0 \stackrel{\text{def}}{=} |\mathcal{N}_0| \gg N$).*

1. *Generate $N$ names uniformly at random (without replacement) from $\mathcal{N}_0$ to form $\mathcal{N}$*
2. *For each attribute $a \in \mathcal{A}$, generate $D$ distinct strings $w_{1,a}, \cdots, w_{D,a} \in \mathcal{T}^L$ uniformly at random (without replacement) to form the diversity set $\mathcal{D}_a$.*
3. *For each name $n \in \mathcal{N}$ and attribute $a \in \mathcal{A}$, generate value $v^\star(n,a) = (v_1, v_2, \cdots, v_C)$ by sampling each $v_i \in \mathcal{D}_a$ uniformly at random.*

*Let $\mathcal{Z} \stackrel{\text{def}}{=} \{(n, a, v^\star(n, a)\}_{n \in \mathcal{N}, a \in \mathcal{A}}$ be the knowledge set.*

**Proposition 2.3** (trivial, bit complexity upper bound). *Given $\mathcal{N}_0$ and $\mathcal{A}$ and $\mathcal{T}$, to describe a knowledge set generated in Definition 2.2, one needs at most the following number of bits:*

$$\log_2 \binom{|\mathcal{N}_0|}{N} + NKC \log_2 D + K \log_2 \binom{T^L}{D} \approx N \log_2 \frac{|\mathcal{N}_0|}{N} + NKC \log_2 D + KD \log_2 \frac{T^L}{D} \ .$$

(The approximation is valid when $|\mathcal{N}_0| \gg N$ and $T^L \gg D$.) We will present a bit complexity lower bound in Section 3.

## 2.2 Knowledge (Empirical Setting)

We utilize both the synthetic bioD dataset, generated as per Definition 2.2, and several human biography datasets to evaluate language model scaling laws.

Allen-Zhu and Li [3] introduced a synthetic biography dataset comprising $N$ individuals, each characterized by six attributes: birth date, birth city, university, major, employer, and working city.[9] To translate these tuples into natural language, in their bioS dataset, each individual is described by six randomly selected English sentence templates corresponding to their attributes. We direct readers to their paper for more details but provide an example below:

> Anya Briar Forger was born on October 2, 1996. She spent her early years in Princeton, NJ. She received mentorship and guidance from faculty members at Massachusetts Institute of Technology. She completed her education with a focus on Communications. She had a professional role at Meta Platforms. She was employed in Menlo Park, CA.

(2.1)

In this paper, we explore three variations of such datasets:

- bioS($N$) represents an online dataset for $N$ individuals, where each biography is generated with new randomness for the *selection* and *ordering* of six sentence templates *on-the-fly*.
- bioS$^{\text{simple}}$($N$) denotes a similar dataset, but here, each biography is generated once with a fixed random selection and ordering of the sentence templates.
- bioR($N$) refers to the same dataset, but with each biography written 40 times by LLaMA2 [33] to increase realism and diversity.

These datasets correspond to the bioS multi+permute, bioS single+permute, and bioR multi data types discussed in [3], albeit with minor differences. While their study focused on $N = 100K$, we expand our scope for bioS to consider $N$ up to $20M$; for bioR, we limit $N$ to $1M$, which already yields a dataset size of 22GB.

As introduced in Section 1, if each knowledge piece is seen 1000 times during training, we call this 1000 exposures. For bioS($N$), 1000 exposures will unlikely include identical biography data because there are 50 sentence templates for each attribute and a total of $50^6 \times 6!$ possible biographies per person. For bioS$^{\text{simple}}(N)$, 1000 exposures mean 1000 passes of the data. For bioR($N$), 1000/100 exposures mean only 25/2.5 passes of the training data.

---

[9] All attributes, except for the working city (determined by the employer's headquarters), are chosen uniformly and independently at random. There are $N_0 = 400 \times 400 \times 1000$ possible person names, $12 \times 28 \times 200$ birth dates, 200 birth cities, 300 universities, 100 majors, and 263 employers. Additionally, a random pronoun with 2 possibilities is chosen for each person.

For the bioD dataset, we define $\mathcal{N}_0$ to be identical to bioS, with $|\mathcal{N}_0| = 400 \times 400 \times 1000$. We encapsulate a person's attributes within a single paragraph, employing random sentence orderings and a consistent sentence template. For example:

Anya Briar Forger's ID 7 is $v_{7,1}, \ldots, v_{7,C}$. Her ID 2 is $v_{2,1}, \ldots, v_{2,C}$. [...] Her ID 5 is $v_{5,1}, \ldots, v_{5,C}$.

In this paper, we primarily utilize bioS. To illustrate broader applicability and *to better connect to theoretical bounds*, we also present results for bioS$^{\text{simple}}$, bioR, and bioD.

## 2.3 Models and Training

GPT2 was introduced in [26]. Due to its limitations from the absolute positional embedding [2], we adopt its modern variant, *rotary positional embedding* [7, 31], which we still refer to as GPT2 for convenience. Additionally, we disable dropout, which has been shown to improve performance in language models [33]. We explore a wide range of model sizes while using a fixed dimension-per-head of 64. The notation GPT2-$\ell$-$h$ represents $\ell$ layers, $h$ heads, and $64h$ dimensions; for example, GPT2-small corresponds to GPT2-12-12. The default GPT2Tokenizer is used, converting people's names and most attributes into tokens of variable lengths. In examining the impact of model architectures on scaling laws in Section 7, we will also use LLaMA/Mistral architectures [19, 32].

**Training.** We train language models *from scratch (i.e., random initialization)* using the specified datasets. Knowledge paragraphs about individuals are randomly concatenated, separated by `<EOS>` tokens, and then randomly segmented into 512-token windows. The standard autoregressive loss is employed for training. Unless specified otherwise, training utilizes the default AdamW optimizer and mixed-precision fp16. Learning rates and weight decays are moderately tuned (see appendix).

## 3 Bit Complexity Lower Bound

When assessing the knowledge stored in a model, we **cannot** simply rely on the **average, word-by-word** cross-entropy loss. For example, the phrase "received mentorship and guidance from faculty members" in (2.1) does not constitute useful knowledge. We should instead focus on the *sum* of the loss for *exactly* the knowledge tokens.

Consider a model $F$ with weight parameters $W \in \mathcal{W}$. Assume $F$ is trained on a bioD($N, K, C, D, L, T$) dataset $\mathcal{Z}$ as defined in Definition 2.2 using any optimizer; this process is represented as $W = W(\mathcal{Z})$ (the model's weight is trained as a function of the training dataset $\mathcal{Z}$). During the evaluation phase, we express $F$ through two functions: $F^\top(W, R)$, which generates names, and $F^\perp(W, n, a, R)$, which generates values given $(n, a)$, where $R$ denotes the randomness used in generation. Let $F_1^\perp(W(\mathcal{Z}), n, a, R)$ represent the first chunk of $F^\perp(W(\mathcal{Z}), n, a, R)$. We evaluate $F$ by calculating the following three cross-entropy losses:[10]

$$\mathbf{loss}_{name}(\mathcal{Z}) \stackrel{\text{def}}{=} \mathop{\mathbb{E}}_{n \in \mathcal{N}} -\log \mathop{\mathbf{Pr}}_{R} \left[ F^\top(W(\mathcal{Z}), R) = n \right]$$

$$\mathbf{loss}_{value1}(\mathcal{Z}) \stackrel{\text{def}}{=} \mathop{\mathbb{E}}_{n \in \mathcal{N}, a \in \mathcal{A}} -\log \mathop{\mathbf{Pr}}_{R} \left[ F_1^\top(W(\mathcal{Z}), n, a, R) = v_1^\star(n, a) \right]$$

$$\mathbf{loss}_{value}(\mathcal{Z}) \stackrel{\text{def}}{=} \mathop{\mathbb{E}}_{n \in \mathcal{N}, a \in \mathcal{A}} -\log \mathop{\mathbf{Pr}}_{R} \left[ F^\perp(W(\mathcal{Z}), n, a, R) = v^\star(n, a) \right]$$

---

[10]We use $\mathbb{E}_n$ or $\mathbb{E}_{n,a}$ to denote uniform random selection of $n \in \mathcal{N}, a \in \mathcal{A}$.

*Remark* 3.1. For a language model, such quantities can be *computed from* its auto-regressive cross-entropy loss. For instance, when evaluating the model on the sentence "Anya Briar Forger's ID 7 is $v_{7,1}, \ldots, v_{7,C}$," *summing up* (not averaging!) the loss over the tokens in "Anya Briar Forger" yields exactly $-\log \mathbf{Pr}_R \left[ F^\top(W(\mathcal{Z}), R) = n \right]$ for $n$ = "Anya Briar Forger"; *summing up* the loss over the token $v_{7,1}$ results in $-\log \mathbf{Pr}_R \left[ F_1^\top(W(\mathcal{Z}), n, a, R) = v_{7,1} \right]$ for this $n$ and $a$ = "ID 7"; and *summing up* the loss over the entire sequence $v_{7,1}, \ldots, v_{7,C}$ gives $-\log \mathbf{Pr}_R \left[ F^\top(W(\mathcal{Z}), n, a, R) = v_{7,1}, \ldots, v_{7,C} \right]$. This holds *regardless* of the tokenizer or value length.

**Theorem 3.2** (bit complexity lower bound). *Suppose $N \geq \Omega(D \log N)$. We have*

$$\log_2 |\mathcal{W}| \geq \mathop{\mathbb{E}}_{\mathcal{Z}} \left[ N \log_2 \frac{N_0 - N}{e^{\mathbf{loss}_{name}(\mathcal{Z})}} + NK \log_2 \frac{D^C}{e^{\mathbf{loss}_{value}(\mathcal{Z})}} + KD \log_2 \frac{T^L - D}{De^{(1+o(1))\mathbf{loss}_{value1}(\mathcal{Z})}} - o(KD) \right]$$

$$= N \log_2 \frac{N_0 - N}{e^{\mathbb{E}_\mathcal{Z} \mathbf{loss}_{name}(\mathcal{Z})}} + NK \log_2 \frac{D^C}{e^{\mathbb{E}_\mathcal{Z} \mathbf{loss}_{value}(\mathcal{Z})}} + KD \log_2 \frac{T^L - D}{De^{(1+o(1))\mathbb{E}_\mathcal{Z} \mathbf{loss}_{value1}(\mathcal{Z})}} - o(KD)$$

The goal of the paper is to study how the number of model parameters competes with this bound.

**Corollary 3.3** (no-error case). *In the ideal case, if for every data $\mathcal{Z}$, $F$ can generate a name from $\mathcal{N}$ with exact $1/N$ probability each, then $\mathbf{loss}_{name}(\mathcal{Z}) = \log N$; and if $F$ can 100% accurately generate values given $(n, a)$ pairs, then $\mathbf{loss}_{value}(\mathcal{Z}) = \mathbf{loss}_{value1}(\mathcal{Z}) = 0$. In such a case,*

$$\log_2 |\mathcal{W}| \geq N \log_2 \frac{N_0 - N}{N} + NKC \log_2 D + KD \log_2 \frac{T^L - D}{D} - o(KD)$$

*asymptotically matching the upper bound Proposition 2.3.*

*Remark* 3.4 (why "sum of 3"). It is essential to obtain a lower bound that is the *sum* of the three components; neglecting any may result in a suboptimal bound (see examples in Appendix A.4).

*Remark* 3.5 (why "random data"). Studying a lower bound for a fixed dataset $\mathcal{Z}$ is impossible — a model could hard-code $\mathcal{Z}$ into its architecture even without any trainable parameter. Therefore, it is necessary to consider a lower bound with respect to a *distribution* over datasets.

**Proof difficulties.** If names are fixed ($\mathcal{N} = \mathcal{N}_0$) and there are $N$ pieces of knowledge, each uniformly chosen from a fixed set $[T]$, it is straightforward that any model $F(W)$, capable of learning such knowledge *perfectly*, must satisfy $\log_2 |\mathcal{W}| \geq N \log_2 T$. To relate this to Theorem 3.2, we encounter three main challenges. First, the model $F$ may only learn the knowledge with a certain degree of accuracy, as defined by the cross-entropy loss. Second, $\mathcal{N} \neq \mathcal{N}_0$ so names need to be learned — even a perfect model cannot achieve zero cross-entropy loss when generating names. Third, there is a dependency between knowledge pieces — the value depends on the name and the choice of the diversity set (i.e., $\mathcal{D}_a$). The proof of Theorem 3.2 is deferred to Appendix F.

## 4 Capacity Ratio

Motivated by Theorem 3.2, ignoring lower order terms, we define the empirical capacity ratio as

**Definition 4.1.** *Given a model $F$ with $P$ parameters trained over a bioD$(N, K, C, D, L, T)$ dataset $\mathcal{Z}$, suppose it gives $p_1 = \mathbf{loss}_{name}(\mathcal{Z})$, $p_2 = \mathbf{loss}_{value}(\mathcal{Z})$, $p_3 = \mathbf{loss}_{value1}(\mathcal{Z})$, we define its capacity ratio and max capacity ratio*

$$R(F) \stackrel{\text{def}}{=} \frac{N \log_2 \frac{N_0}{e^{p_1}} + NK \log_2 \frac{D^C}{e^{p_2}} + KD \log_2 \frac{T^L}{De^{p_3}}}{P}.$$

$$R^{\mathsf{max}}(F) \stackrel{\text{def}}{=} \frac{N \log_2 \frac{N_0}{N} + NKC \log_2 D + KD \log_2 \frac{T^L}{D}}{P}.$$

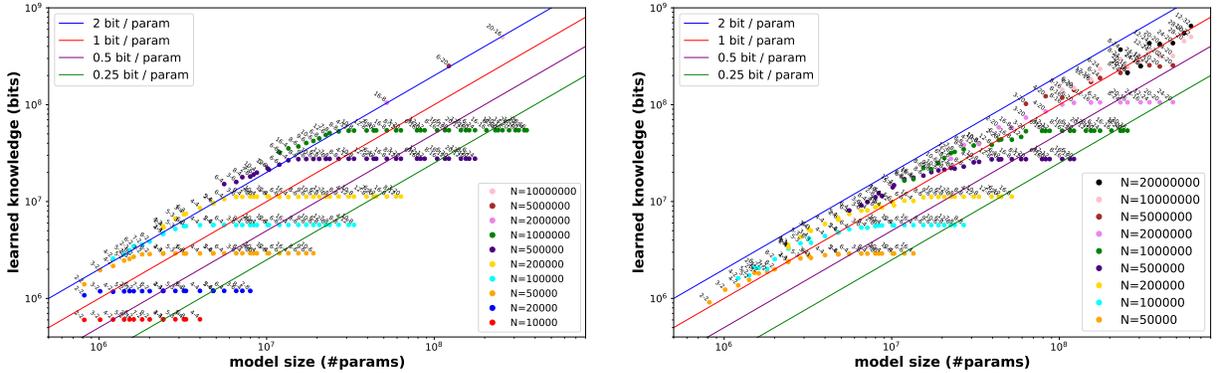

(a) bioS($N$) data — **1000 exposures** — peak $R(F) \geq 2$  (b) bioS($N$) data — **100 exposures** — peak $R(F) \geq 1$

Figure 1: Scaling laws for GPT2 pretrained on bioS($N$) data using fp16 (mixed-precision) for 1000/100 exposures.

---

**Conclusion.** The *peak* capacity ratios consistently exceed $R(F) \geq 2$ (resp. $\geq 1$) for 1000 exposures (resp. 100 exposures) of pretraining on each knowledge piece, **regardless of model depth/size**.

---

**Remarks.** Each dot $\ell$-$h$ represents GPT2 with $\ell$ layers, $h$ heads, and $64d$ dimensions. The learned knowledge is calculated by the bit-complexity lower bound Theorem 3.2. The appendix also includes: Figure 11 showing similar results for bioS$^{\mathsf{simple}}$($N$) and bioR($N$) data, Figure 14 demonstrating that the *same holds* for quantization using int8, Figure 10 confirming full extractability of all learned knowledge.[13]

---

**Larger models?** Training GPT2-20-16 on bioS($10M$) for 1000 exposures costs 8.5 days with 64 A100s, while GPT2-12-32 on bioS($20M$) for 100 exposures took 2.4 days. In our synthetic setting, we see no need to scale up further. Instead, we prefer to allocate GPUs to explore other aspects covered in this paper.

---

*Remark 4.2.* One must have $R(F) \leq R^{\mathsf{max}}(F)$, and equality is obtained if the model is *perfect*. For a fixed dataset, further increases in model size do not yield additional knowledge, thus $R^{\mathsf{max}}(F)$ approaches zero as the model size $P$ increases. On the other hand, Theorem 3.2 implies, ignoring lower-order terms, that if the model parameters are 8-bit (such as int8), then $R(F) \leq 8$.

For our bioS($N$) data, we define a slightly reduced capacity ratio by omitting the diversity term.[11]

**Definition 4.3.** *Given a model $F$ with $P$ parameters trained over the bioS($N$) dataset $\mathcal{Z}$, suppose it gives $p_1 = \mathbf{loss}_{name}(\mathcal{Z})$ and $p_2 = \mathbf{loss}_{value}(\mathcal{Z})$, its capacity ratio*[12]

$$R(F) \stackrel{\text{def}}{=} \frac{N \log_2 \frac{N_0}{e^{p_1}} + N \log_2 \frac{S_0}{e^{p_2}}}{P} \quad \text{and} \quad R^{\mathsf{max}}(F) \stackrel{\text{def}}{=} \frac{N \log_2 \frac{N_0}{N} + N \log_2 S_0}{P}$$

*for $N_0 = 400 \times 400 \times 1000$ and $S_0 = 2 \times (12 \cdot 28 \cdot 200) \times 200 \times 300 \times 100 \times 263$ (c.f. Footnote 9).*

*Remark 4.4.* Ignoring names, each person contains $\log_2(S_0) \approx 47.6$ bits of knowledge.

## 5 Base Scaling Laws

---

[11] A version of Theorem 3.2 can be proven for this dataset with a simpler proof, as it excludes the diversity set. This could also mean the model has full prior knowledge of the diversity set (e.g., assuming a fixed set of 300 university names) without counting this knowledge towards its learned bits.

[12] Here, one can let $\mathcal{K} = \{\text{birth date}, \text{birth city}, \text{university}, \text{major}, \text{employer}, \text{gender}\}$ and accordingly define $\mathbf{loss}_{value}(\mathcal{Z}) \stackrel{\text{def}}{=} \mathbb{E}_{n \in \mathcal{N}} \sum_{a \in \mathcal{K}} -\log \mathbf{Pr}_R \left[ F^{\perp}(W(\mathcal{Z}), n, a, R) = v^{\star}(n, a) \right]$.

[13] A distinction exists between memorizable knowledge (e.g., text memorized during pretraining) and knowledge flexibly extractable via instruction fine-tuning [3]; our results in this paper apply to both.

We first train a series of GPT2 models on the bioS($N$) datasets (see Section 2.2) using mixed-precision fp16. The training protocol ensures that each piece of knowledge is presented 1000 times, a process we refer to as "1000 exposures." It's important to clarify that this differs from making 1000 passes over the data. For example, a single pass through Wiki data might expose the knowledge (US, capital, Washington D.C.) 1000 times, whereas a pass through the Common Crawl might do so a million times. Our synthetic bioS($N$) data, trained for 1000 exposures, aims to replicate such scenarios.[14] Our initial findings are as follows:[15]

> **Result 1** (Figure 1(a)). When trained for 1000 exposures on bioS($N$), with $N$ ranging from 10K to 10M, GPT2 models with sizes from 1M to 0.5B parameters (*irrespective of depth or width*) demonstrate the following:
> 
> (a) the peak capacity ratio $R(F)$ consistently exceeds $R(F) \geq 2$;
> 
> (b) models with $R^{\max}(F) \leq 1.8$ attain near-perfect knowledge accuracies, i.e., $R^{\max}(F) \approx R(F)$;
> 
> (c) across all models, $R(F) \leq 2.3$.

*Remark 5.1.* Result 1(a), 1(b), and 1(c) elucidate *three distinct facets* of the scaling law.

- Result 1(a) highlights the maximum capacity across models; however, this could be misleading if only a single model achieves this peak.
- Result 1(b) reinforces this by showing that all models with a maximum capacity $R^{\max}(F) \leq 1.8$ can achieve such maximum capacity, i.e., $R(F) \approx R^{\max}(F)$. In words, this indicates that for a dataset containing $B$ bits of knowledge, selecting a model size $P \geq B/1.8$ is sufficient.
- Result 1(c) further strengthens this by indicating that no model exceeds capacity ratio 2.3.

***For clarity***, in subsequent results of this paper, we focus solely on the *peak* capacity ratio, with the understanding that observations similar to Result 1(b) and Result 1(c) **consistently apply**.

**Knowledge extraction.** The "2bit/param" result is not about word-by-word memorization. Even better, such knowledge is also flexibly extractable (e.g., via fine-tuning using QAs like "What is Anya Forger's birthday?") [3] and thus can be further manipulated in downstream tasks (such as comparing the birthdates of two people, or performing calculations on the retrieved knowledge, etc.) [4]. This is because our bioS($N$) data is knowledge-augmented: the English biographies have sufficient text diversities [3]. We also verify in Appendix A.2 that such knowledge is extractable.

### 5.1 Data Formats — Diversity and Rewriting

We conduct the same analysis on bioS$^{\mathsf{simple}}$ and bioR. Recall from Section 2.2, bioS$^{\mathsf{simple}}$ is a variant of bioS with reduced text diversity (one biography per person), while bioR is generated by LLaMA2, resulting in close-to-real human biographies. We have:

> **Result 2** (Figure 11 in Appendix A.3). In the same 1000-exposure setting, peak capacity ratios for GPT2 trained on bioS$^{\mathsf{simple}}$ and bioR are also approximately 2, albeit slightly lower. Thus:
> 
> - Diverse data (rewriting the same data multiple times) does not hurt — and may sometimes improve — the model's capacity!

---

[14] Within 1000 exposures, it's likely that the same individual will have 1000 different biography paragraphs detailing the same knowledge (see Section 2.2). Therefore, 1000 exposures can occur within a single pass.

[15] We here focus on GPT2 models with depth $\geq 2$, and 1-layer models show slightly lower capacity ratios (see Figure 9). Our model selection covers most natural combinations of transformer width/depth, details in Appendix A.

Let's highlight the significance of Result 2. Recall from Section 2.2:

- Training on bioS$^{\mathsf{simple}}$ data for 1000 exposures equals 1000 passes over the data.
- Training on bioS data for 1000 exposures is less than 1 pass.
- Training on bioR data for 1000 exposures equals 25 passes.

Therefore, comparing bioS and bioS$^{\mathsf{simple}}$, it's more advantageous to rewrite the data 1000 times (in this ideal setting), training each for one pass (as done in the bioS data), rather than training the same data for 1000 passes (as done in the bioS$^{\mathsf{simple}}$ data). This is because, without data diversity, the model wastes capacity memorizing sentence structures, resulting in a capacity loss.

In a realistic scenario, tools like LLaMA2 can rewrite pretrain data like we did in bioR. Rewriting data 40 times can produce 40 distinct English paragraphs, sometimes with (different) hallucinated contents. Does this require the model to be 40x larger? No, our comparison between bioS and bioR shows that, if trained for the same duration (40 rewrites each for 25 passes), the model's capacity ratio remains nearly the same, slightly lower due to irrelevant data introduced by LLaMA2.

Allen-Zhu and Li [3] suggested that rewriting pretraining data is crucial for making knowledge extractable rather word-by-word memorization.[16] However, they did not explore the impact on the model's capacity. Our paper addresses this gap, indicating that rewriting pretraining data does not compromise — and may even enhance — the model's knowledge capacity.

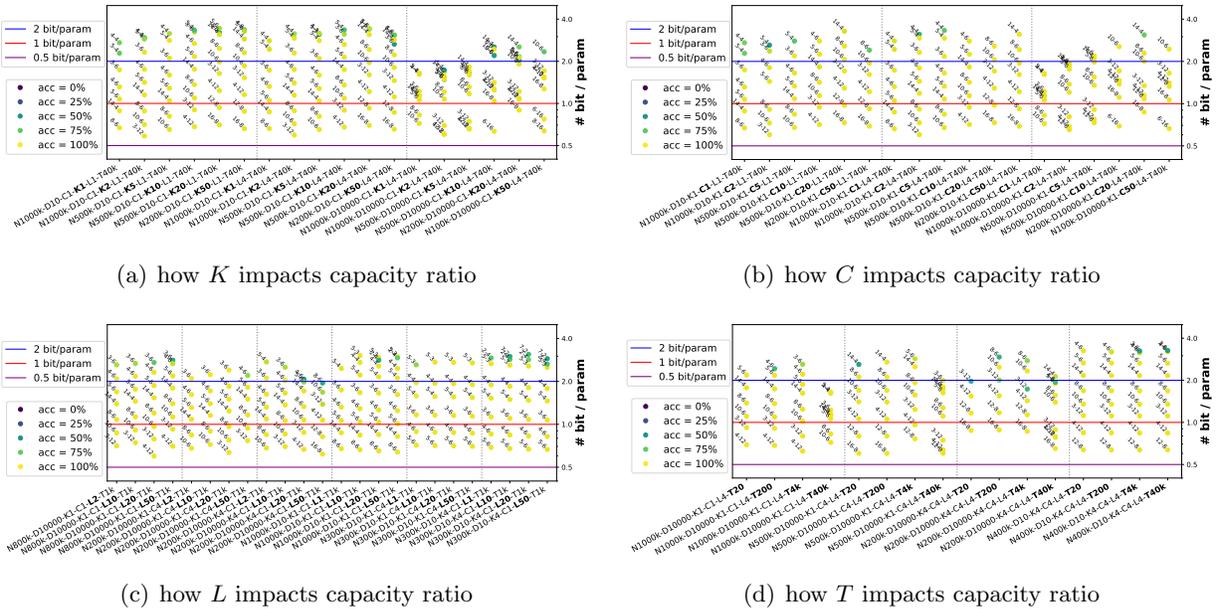

(a) how $K$ impacts capacity ratio  (b) how $C$ impacts capacity ratio

(c) how $L$ impacts capacity ratio  (d) how $T$ impacts capacity ratio

Figure 2: Scaling laws for GPT2 models trained on the bioD$(N, K, C, D, L, T)$ data for **1000 exposures**.

**Conclusion.** The *peak* capacity ratios consistently exceed $R(F) \geq 2$ with a wide range of $K, C, D, L, T$.

*Remarks.* Models with accuracies $\leq 50\%$ are excluded here but included in Figure 12. We disregard $N$'s influence, akin to Figure 1(a), and concentrate on the five hyperparameters $K, C, L, T, D$. Each of the four sub-figures varies a primary hyperparameter while fixing the other four. More details in Appendix A.4.

---

[16] As demonstrated by [3], in low-diversity datasets like bioS$^{\mathsf{simple}}$, knowledge can be word-by-word memorized but is nearly 0% extractable for downstream tasks. Others discover that rewriting data can improve the reversal extractability of knowledge [4, 11].

## 5.2 Parameterized Scaling Laws

We further investigate scaling laws within the $\mathsf{bioD}(N, K, C, D, L, T)$ data family. Unlike with human biographies, where variation is limited to $N$, the $\mathsf{bioD}$ dataset allows for more flexible manipulation of the remaining hyperparameters $K, C, D, L, T$. This enables us to examine how variations in these parameters affect the model's peak capacity.

> **Result 3** (Figure 2). Across a broad spectrum of values, with $K, C$ ranging from 1 to 50, $D$ from 10 to 10,000, $L$ from 1 to 50, and $T$ from 20 to 40,000, we observe that:
> - GPT2 models consistently exhibit a peak capacity ratio $R(F) \geq 2$.

# 6 Training Time vs Scaling Law

What if the model is not sufficiently trained? For instance, there might be instances where knowledge appears only 100 times throughout the pretraining phase. We also calculate the capacity ratios for models trained with 100 exposures on $\mathsf{bioS}(N)$. Our findings can be summarized as follows:

> **Result 4** (Figure 1(b)). When trained for only 100 exposures on the $\mathsf{bioS}(N)$ dataset, with $N$ ranging from 10K to 10M, across a broad spectrum of GPT2 models with sizes from 1M to 0.5B, the peak capacity ratio $R(F)$ consistently exceeds $R(F) \geq 1$.

Therefore, although 1000 exposures may be necessary for a model to reach its maximum storage capacity, training with just 100 exposures results in a capacity loss of no more than 2x.

In Section 10, we shall also consider knowledge that has *extremely low (e.g., 1) or high (e.g., 1M) exposures*. It may not be interesting to study them in isolation, but it becomes more intriguing when they are examined alongside "standard" knowledge, which has appeared, for instance, for 100 exposures, and how this impacts the model's capacity. These will be our Result 10 through 12.

# 7 Model Architecture vs Scaling Law

Several transformer architectures have been widely adopted, with LLaMA and Mistral among the most notable. We outline their key distinctions from GPT2, with further details in Appendix B:
1. LLaMA/Mistral use so-called GatedMLP layers, which is $V(\sigma(W_1 x) \cdot (W_2 x))$ instead of $V\sigma(Wx)$. Shazeer [29] suggested that gated activation might yield marginally improved performance.
2. Unlike GPT2, LLaMA/Mistral do not tie weights.
3. Mistral features larger MLP layers compared to GPT2/LLaMA.
4. Mistral promotes group-query attention, not so by GPT2/LLaMA.
5. LLaMA/Mistral employ a different tokenizer than GPT2.
6. GPT2 uses the *gelu* activation function, LLaMA/Mistral opt for *silu*.
7. GPT2 implements layer normalization with a trainable bias.

Do these architectural variations impact the models' maximum capacities? Our findings suggest that, in terms of knowledge capacity, GPT2 — when enhanced with rotary embedding and without dropout — performs no worse than any other architecture choice above in the sufficient training regime. We summarize the main findings below, deferring details to Appendix B.1:

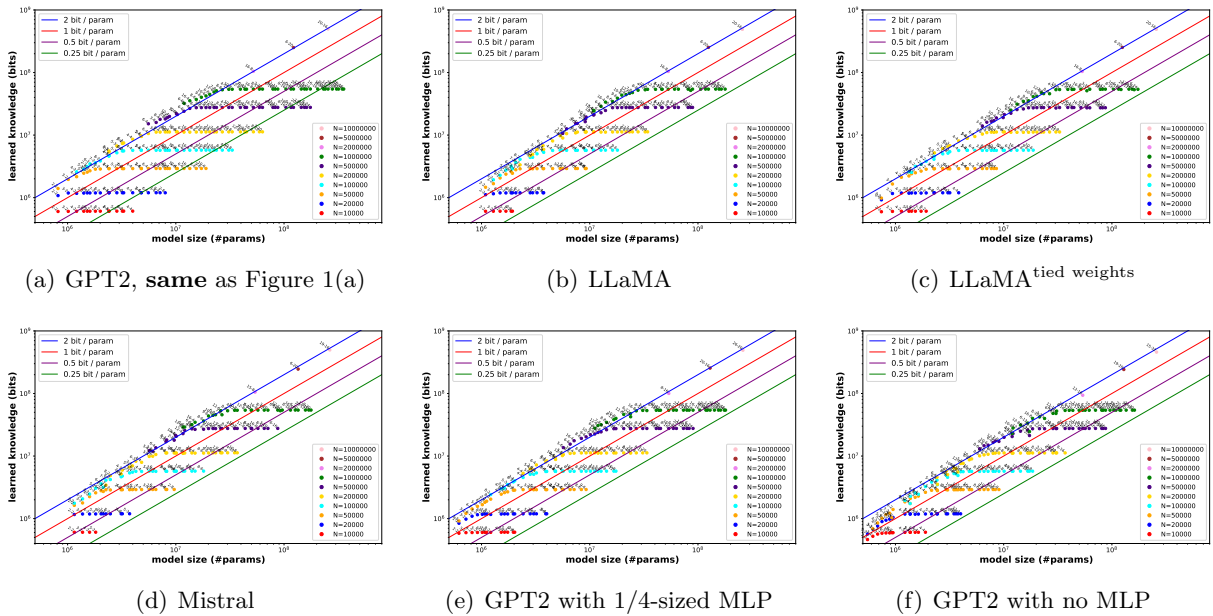

(a) GPT2, **same** as Figure 1(a)   (b) LLaMA   (c) LLaMA$^{\text{tied weights}}$

(d) Mistral   (e) GPT2 with 1/4-sized MLP   (f) GPT2 with no MLP

Figure 3: Scaling laws for other model architectures on the bioS($N$) data with **1000 exposures**.

> **Conclusion.** In the 1000-exposure setting, all model architectures closely follow GPT2's scaling law — including LLaMA/Mistral or even removing the MLP layer completely. The only *very minor* difference is observed in tiny models, where tying the model's (embedding + output layer) weights enhances its capacity, evident from comparing Figure 3(c) with Figure 3(b).

**Result 5** (Figure 3). In the 1000-exposure setting, architectures do not matter much:
- LLaMA architecture performs comparably to GPT2, albeit slightly inferior for the tiny model (i.e., $< 10M$). This discrepancy can be mitigated by also requiring LLaMA architecture to tie weights, as shown in Figure 3(c) compared to Figure 3(b).
- A similar observation applies to Mistral architecture (see Figure 3(d)).
- Reducing the MLP size of GPT2 architecture by 1/4 or even eliminating all MLP layers *does not affect* its capacity ratio, see Figure 3(e) and Figure 3(f). This suggests, contrary to conventional beliefs, the Attention layers are also capable of storing knowledge.

This indicates that the 2bit/param capacity ratio is a relatively *universal law* among most typical (decoder-only) language model architectures.

## 7.1 Insufficient Training Regime and a Closer Comparison

However, differences in architectures become apparent in the insufficient training regime:

**Result 6** (Figure 4). In the 100-exposure setting:
- *Even for large models*, LLaMA architecture's capacity ratio can be 1.3x worse than GPT2, even after optimally tuning learning rates. The results are similar for Mistral.
- Reducing GPT2's MLP size by 1/4 has a negligible impact on the capacity ratio.
- Removing MLPs decreases the capacity ratio by more than 1.5x.

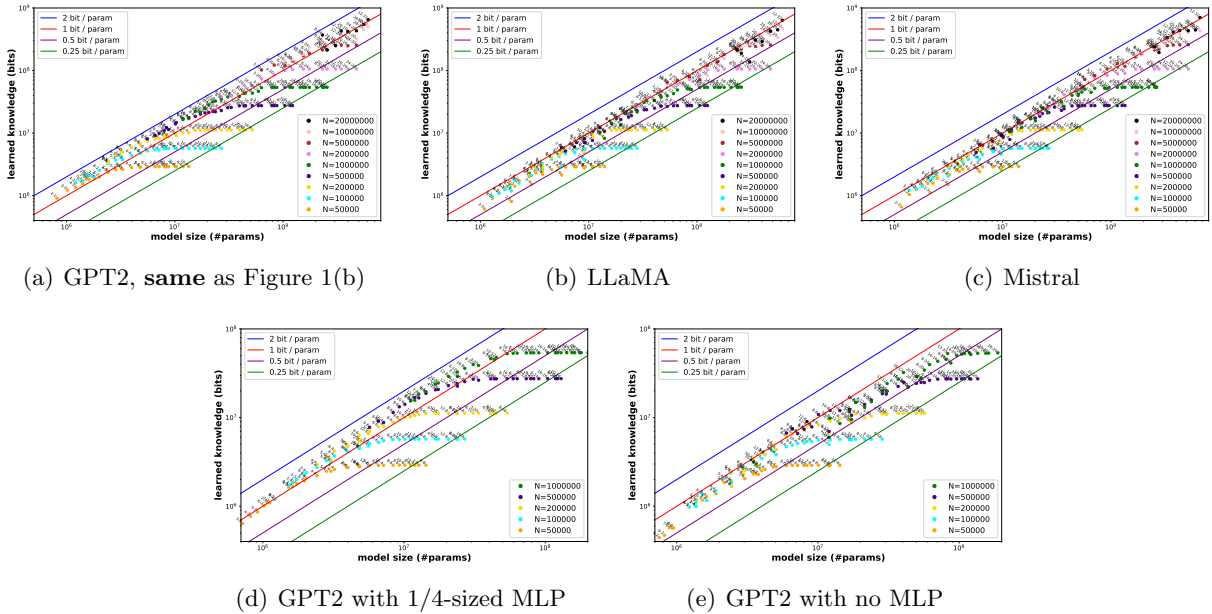

(a) GPT2, **same** as Figure 1(b)      (b) LLaMA      (c) Mistral

(d) GPT2 with 1/4-sized MLP      (e) GPT2 with no MLP

Figure 4: Scaling laws for other model architectures on the bioS$(N)$ data with **100 exposures**.

> **Conclusion.** In the 100-exposure setting, LLaMA/Mistral architectures may underperform GPT2's scaling law by 1.3x, *even for large models*. Reducing the size of GPT2's MLP layer by 1/4 does not affect its scaling law, but removing all MLP layers degrades performance. See Figure 5 for a closer comparison.

To investigate *why* the LLaMA architecture is inferior to GPT2 in the 100-exposure (insufficiently trained) setting, we closely examine LLaMA by gradually modifying its architecture *back towards* GPT2 to identify the key architectural changes. We start by tying weights, as this enhances tiny LLaMA model's capacity in the 1000-exposure setting (Result 5). As illustrated in Figure 5:

- For large models, replacing LLaMA architecture's gated MLP with a standard MLP (while keeping *silu* unchanged) noticeably improves LLaMA's capacity ratio.[17]
- For tiny LLaMA models, switching back to the GPT2Tokenizer is also necessary to match GPT2's performance, though this is a minor issue.[18]
- Other modifications, such as changing from *silu* to *gelu* or adding trainable biases to layer-norms, do not noticeably affect the capacity ratios (so we ignore those figures).

In summary,

> **Result 7.** In the *insufficient training regime* (notably, the 100-exposure setting), except for tiny models, architectural differences generally do not affect performance, except
> - Using gated MLP reduces the model's capacity ratio (Figure 5);
> - Removing all MLP layers lowers the model's capacity ratio, although significantly reducing the size of MLPs (e.g., by a 1/4 factor) does not.

We propose that our experiments with the controllable biography dataset could serve as a valuable testbed for future architectural designs.

---

[17] As discussed in Appendix B, gated MLP layers are less stable to train, thus requiring more time.

[18] This only applies to tiny models and is specific to the biography data we consider here: GPT2Tokenizer may tokenize years such as 1991 into a single token, while LLaMATokenizer will tokenize it into four digit tokens.

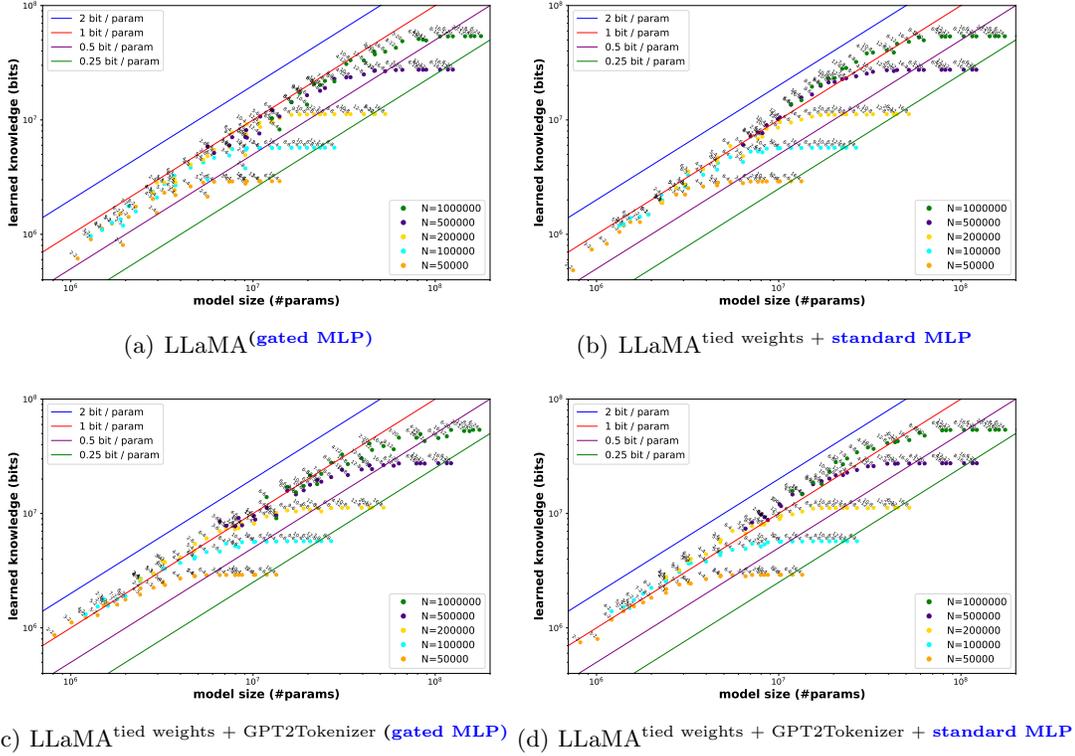

(a) LLaMA$^{\text{(gated MLP)}}$  (b) LLaMA$^{\text{tied weights}}$ + **standard MLP**

(c) LLaMA$^{\text{tied weights}}$ + GPT2Tokenizer (**gated MLP**)  (d) LLaMA$^{\text{tied weights}}$ + GPT2Tokenizer + **standard MLP**

Figure 5: A closer comparison on LLaMA's scaling laws with bioS($N$) data for **100 exposures**.

**Conclusion.** Switching from gated MLP to a standard MLP (*left vs. right*) enhances larger model's capacity ratios. For tiny models, using GPT2Tokenizer (*top vs. bottom*) is beneficial (this is a minor point).

**Remarks.** For a strong comparison we used one learning rate choice in Figure 5(d) and 5(b), but present the best among three choices for Figure 5(a) and 5(c). Further details can be found in Appendix B.2.

## 8 Quantization vs Scaling Laws

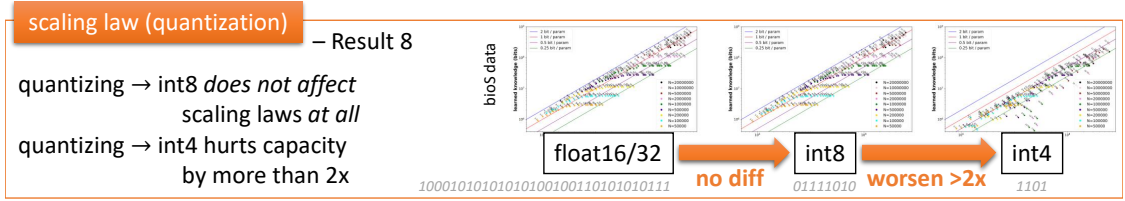

Figure 6: Illustration of Result 8 (see Figure 14 for bioS data and Figure 15 for bioD data in Appendix C).

We have trained and tested models using (mixed precision) 16-bit floats. What happens if we quantize them to int8/int4 after training? We used the `auto_gptq` package, which is inspired by the GPTQ paper [10], for quantization.

**Result 8** (Figure 6). Quantizing language models (e.g., GPT2) trained with 16-bit floats:
- to int8 has a negligible impact on their capacity ;
- to int4 reduces their capacity by more than 2x.

Thus, **even for** models at peak capacity of 2 bits/param, quantizing to int8 does not affect capacity. Given that 2 bits/param was the best capacity ratio even after 1,000 training exposures on high-quality data, we conclude that extending training *may not* further improve the model's capacity, *but quantization can.*

Since an int8-based model has an absolute upper bound $R(F) \leq 8$ on capacity ratio, we have:

**Corollary 8.1.** *Language models, like GPT2, can exceed 1/4 of the absolute theoretical limit for storing knowledge.*

Unfortunately, using this quantization package, reducing the model to int4 significantly diminishes its capacity (more than 2x loss from int8 to int4). This suggests for high-quality int4 models, incorporating quantization during training may be necessary.

### 8.1 Where Is the Knowledge Stored?

We have seen that LLMs can efficiently compress knowledge into their parameter space, achieving 2bit/param even with 8-bit parameters. This raises the question: how and where is such knowledge stored? Our preliminary answer is that knowledge can be compactly stored within the model in a not-so-redundant manner. It is unlikely that the MLP layers alone store knowledge, as Attention layers, being of comparable sizes, also contribute to knowledge storage (c.f. Result 5). Moreover, particularly in models near the capacity boundary, removing the last transformer layer of an $L$-layer model to "probe" for remaining knowledge reveals that the "leftover knowledge" can be significantly less than $1 - \frac{1}{L}$ of the total.[19] This suggests knowledge is stored not in individual layers but in a complex manner, akin to a safe with combination locks, where removing one layer may eliminate much more than $\frac{1}{L}$ of the total knowledge.

## 9 Mixture of Experts vs Scaling Laws

An important way to enhance efficiency in modern language models is the incorporation of sparsity. The Mixture of Experts (MoE) architecture plays a crucial role in this regard [9, 30]. A question arises: does the MoE model scale differently in terms of the capacity ratio? For an MoE model, let $P$ denote the **total number of parameters in the model**, including all experts. Due to its inherent sparsity, the effective number of parameters can be significantly less than $P$. Our primary observation is that MoE models scale *similarly* to dense models, even with 32 experts per layer.

Consider, for instance, GPT2, but with its MLP layer ($d \to 4d \to d$) replaced by 32 experts, each following a $d \to d \to d$ configuration. This setup uses $64d^2$ total parameters, but during inference, only $2d^2$ parameters are used per token (e.g., when using $topk = 1$). After including the Attention layers, which each have $4d^2$ parameters, the ratio between the total and the effective number of parameters for the 32-expert MoE models is approximately $\frac{4d^2 + 64d^2}{4d^2 + 2d^2} \approx 11.3$.

One might wonder, given that during inference time, the model uses only 11.3x fewer parameters, whether this affects the model's capacity ratio by a factor close to 11.3x or closer to 1x? We show:

---

[19]This experiment, deemed not particularly interesting, was omitted from the paper. The probing technique used is Q-probing from [3].

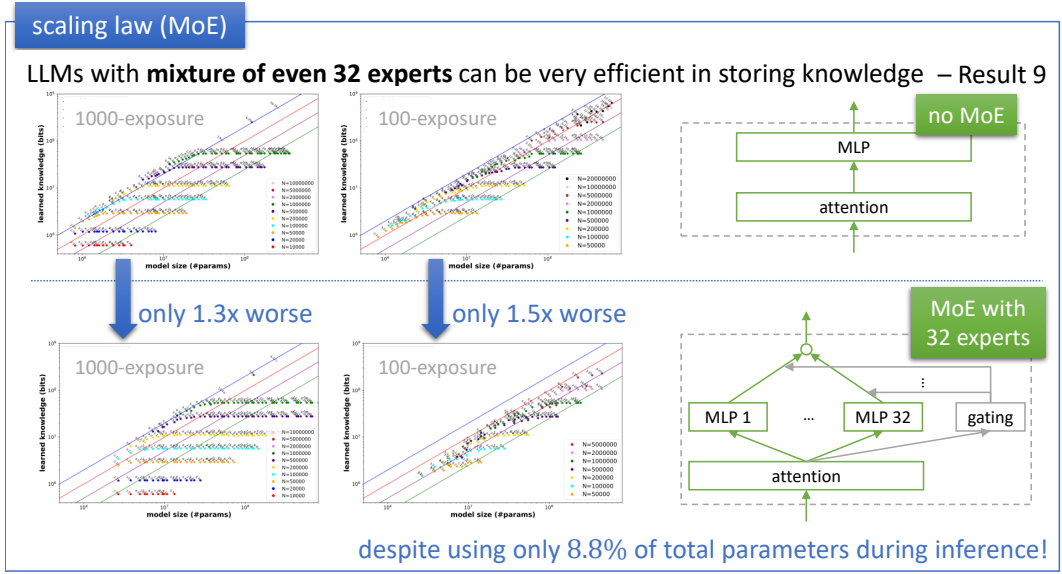

Figure 7: Illustration of Result 9 (details see Figure 16 in Appendix D).

**Result 9** (Figure 7). MoE is nearly fully efficient in storing knowledge, capable of leveraging all its parameters despite the sparsity constraint.

Specifically, consider the GPT2-MoE model with 32 experts. If we compute its capacity ratio with respect to the *total* number of parameters and compare that to GPT2:

- in the 1000-exposure settings, the peak capacity ratio decreases by 1.3x; and
- in the 100-exposure settings, the peak capacity ratio decreases by 1.5x.

*Remark* 9.1 (topk). Result 9 holds even in the "sparsest" setting where $topk = 1$ and $cap\_factor = 2$ in the MoE routing. The results are similar when using $topk = 2$ and $cap\_factor = 1$ or $topk = 2$ and $cap\_factor = 2$— we discuss more in Appendix D.

*Remark* 9.2. It is typically observed in practice that MoE models underperform compared to dense models with the same number of total parameters. We demonstrate that this degradation does not come from the model's knowledge storage capability.

## 10 Junk Data vs Scaling Laws

Not all data are useful for knowledge acquisition. For instance, while Wikipedia is full of valuable information, the Common Crawl of web pages may not be (there are also many pieces of information on those webpages, but they may not be useful for a language model to learn, such as the serial number of a random product). How does the presence of low-quality data impact the scaling laws of *useful knowledge capacity*? To investigate this, we create a mixed dataset where:

- 1/8 of tokens originate from $\mathsf{bioS}(N)$ for various $N$ (referred to as *useful data*), and
- 7/8 of tokens originate from $\mathsf{bioS}(N')$ for a large $N' = 100M$ (referred to as *junk data*).

We train models on this mixture, ensuring each piece of useful data is seen for 100 exposures, thus making the total training 8 times longer compared to 100 exposures without junk (i.e., Figure 1(b)). We focus on the capacity ratio of the useful data (the data in $\mathsf{bioS}(N)$) and compare that to

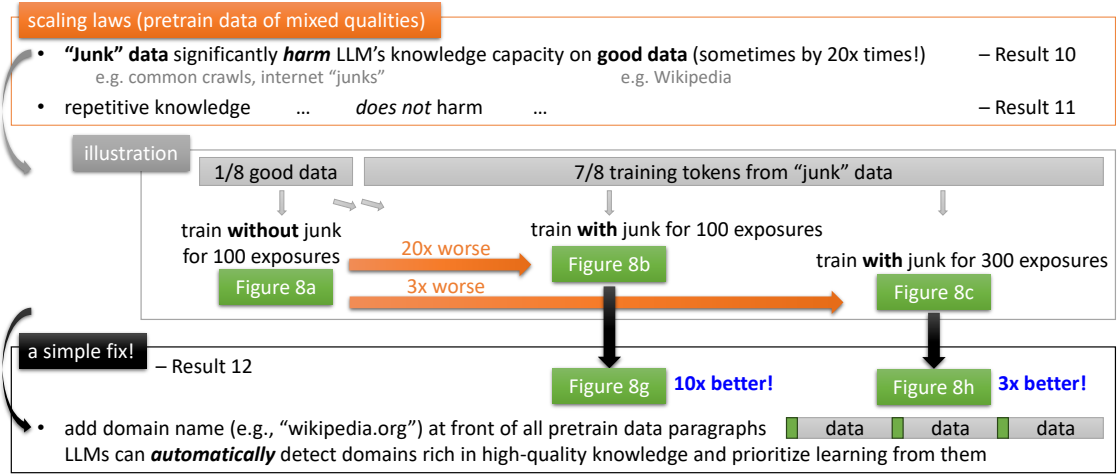

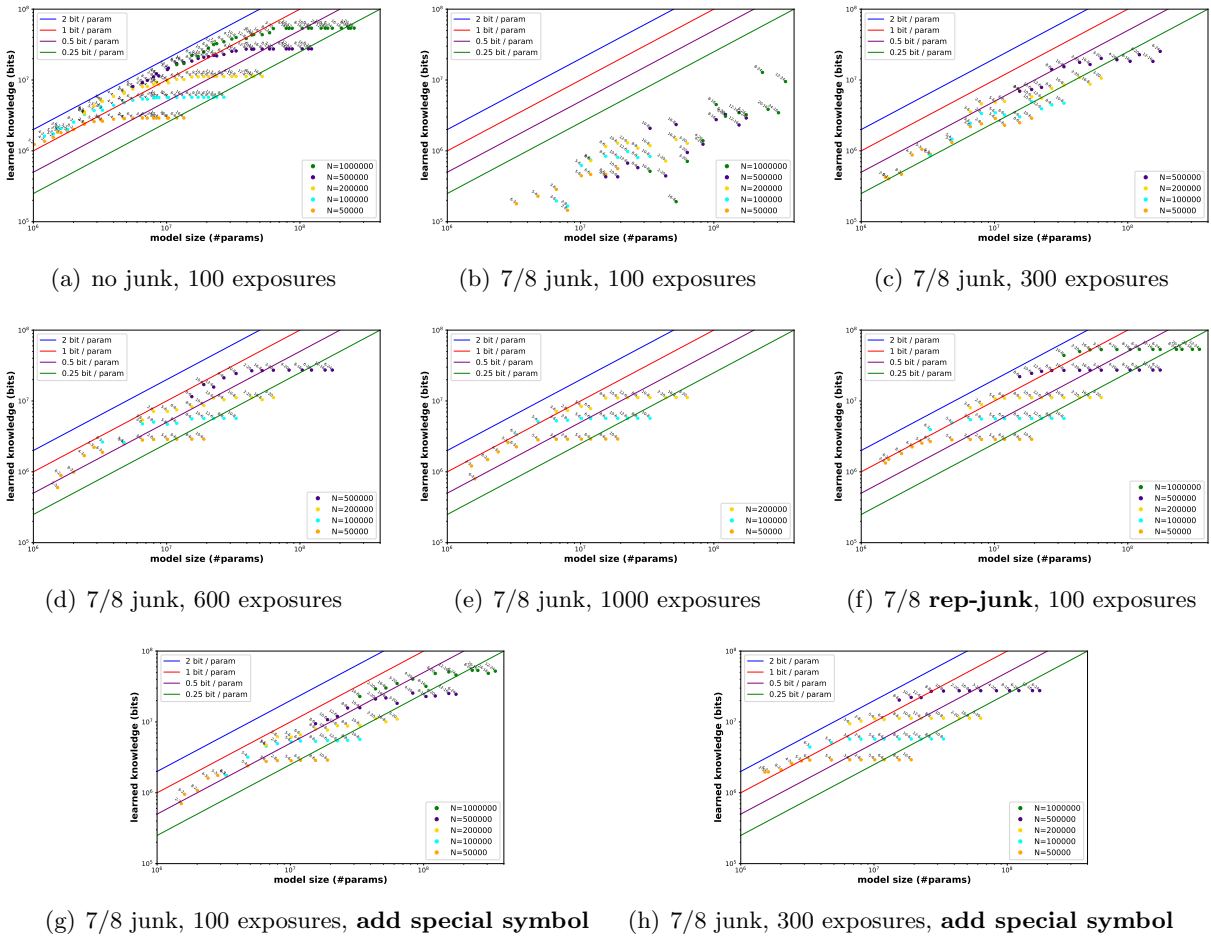

Figure 8: Capacity ratios with 7/8 junk data (useful data observed 100/300/600/1000 exposures during pretraining).

> **Conclusions.** In Figure 8(b)-8(e), when junk data mimics random knowledge, capacity ratios are *significantly impacted* unless training time is substantially increased. In Figure 8(f), if the junk data is highly repetitive, there is no degradation. In Figure 8(g)+8(h), *adding a special symbol token* to useful data, akin to domain names like wikipedia.org, *mitigates capacity degradation*.

> *Remarks.* Training for Figure 8(e) takes 80x longer than for Figure 8(a), making experiments with larger data sizes $N$ expensive (but anyways unnecessary).

Figure 1(b).[20] How much does the capacity ratio degrade in the presence of junk data?

> **Result 10** (Figure 8(a)-8(e))**.** When 7/8 of the training tokens come from junk data (i.e., bioS($N'$) for $N' = 100M$), transformer's learning speed for useful data significantly degrades:
> - If trained for the same 100 exposures, the capacity ratio may degrade by 20x compared with training without junk (compare Figure 8(b) with Figure 8(a)).
> - Even trained for 300/600/1000 exposures, the capacity ratio still degrades by 3x/1.5x/1.3x compared with 100 exposures without junk (Figure 8(c), 8(d), and 8(e) vs. Figure 8(a)).

This underscores the *crucial importance of pretrain data quality*: even if junk data is entirely random, it negatively impacts model's knowledge capacity even with sufficient training.

In contrast, if 7/8 of data is bioS($N'$) with a very small $N'$, simulating highly repetitive knowledge appearing in training tokens (e.g., "da Vinci painted the Mona Lisa" in millions of variations), this may not affect the model's capacity for "standard" knowledge (e.g., those with 100 exposures):

> **Result 11** (Figure 8(f))**.** If 7/8 of the training tokens come from highly repetitive data (i.e., bioS($N'$) for $N' = 1K$), this does not affect the learning speed of useful knowledge:
> - The 100-exposure capacity ratio of useful data is unchanged (Figure 8(f) vs. Figure 8(a)).

Finally, if pretrain data's quality is poor and hard to improve, **a backup strategy exists:**

> **Result 12** (Figure 8(g)+8(h))**.** When 7/8 of training tokens are from junk (i.e., bioS($N'$) for $N' = 100M$), adding a special token at the start of every useful data greatly improves capacity ratio:
> - With 100 exposures, the capacity ratio degrades only by 2x (Figure 8(g) vs. Figure 8(a)).
> - With 300 exposures, the capacity ratio matches that of the 100-exposure scaling law without junk (compare Figure 8(h) with Figure 8(a)).

Let us connect Result 12 to practice. First, adding a special token to high-credibility data is very practical: imagine adding the domain name "wikipedia.org" at the beginning of all Wikipedia paragraphs. (Adding a special token to junk data would be less meaningful.)[21]

More generally, one can envision adding domain names (e.g., wikipedia.org) to every piece of the pretraining data. This would significantly enhance the model's knowledge capacities, because Result 12 demonstrates that **language models can automatically detect which domains are rich in high-quality knowledge and prioritize learning from them**. We emphasize that the model does not need any prior knowledge to identify which domains contain high-quality knowledge; **this process is entirely autonomous**.

## 11 Conclusion

We investigated the scaling laws of language models, specifically the relationship between model size and the total bits of knowledge stored. Our findings reveal a *precise, universal* scaling law: a sufficiently-trained transformer (i.e., one whose training loss has plateau-ed) can store 2 bits of knowledge per parameter, even when quantized to int8, which is only 1/4 away from the information-theoretical maximum. We also examined how these scaling laws are influenced by various hy-perparameters, including training duration, model architectures, floating-point precision, sparsity constraints like MoE, and data signal-noise ratios.

---

[20]The model's ability to learn from junk data is negligible; each person in bioS($N'$) appears only 0.2 times during training when $N = 200k$, or 0.05 times when $N = 50k$.

[21]Result 12 also holds if one adds a (unique) special token for every piece of junk data; however, this could be meaningless as junk data often originates from various websites, making it hard to assign a unique identifier.

In terms of knowledge capacity, our methodology provides a **more accurate and principled playground** for comparing model architectures, training techniques, and data quality. We believe this playground can assist practitioners in making informed decisions about model selection, training data preparation, and further theoretical research into LLMs. Finally, our research represents an initial step towards addressing a fundamental question: how large does a language model need to be? We hope our findings will inspire further research in this area. Ultimately, we aim to provide a principled answer to the question, "Are language models with 1T parameters sufficient to achieve AGI?" in the future.

# Appendix

## A More on GPT2 Scaling Laws

In this paper, our primary focus is on bioS($N$) for $N$ ranging between 10K and 20M. Notably, bioS($20M$) encompasses approximately 1B bits of knowledge (refer to Theorem 3.2).

**GPT2 model.** As elaborated in Section 2.3, we refer to the original GPT2 model [26] as GPT2, *after* substituting its positional embedding with *rotary embedding* [7, 31] and removing its dropout layer [33]. These modifications are widely recognized for enhancing performance in language modeling tasks (see also [2] for a controlled experiment comparing that). We explore various GPT2 model sizes, maintaining a dimension-per-head of 64. The notation GPT2-$\ell$-$h$ represents the (modified) GPT2 architecture with $\ell$ layers, $h$ heads, and $64h$ dimensions. The context length is set to 512.

Details on our specifications of LLaMA, Mistral, and other architectures will be provided in Appendix B as needed.

**Model sizes.** In this study, we calculate model sizes *after excluding* all unused tokens in the embedding layer. For example, while the GPT2 embedding layer typically has $50256 \times (64h)$

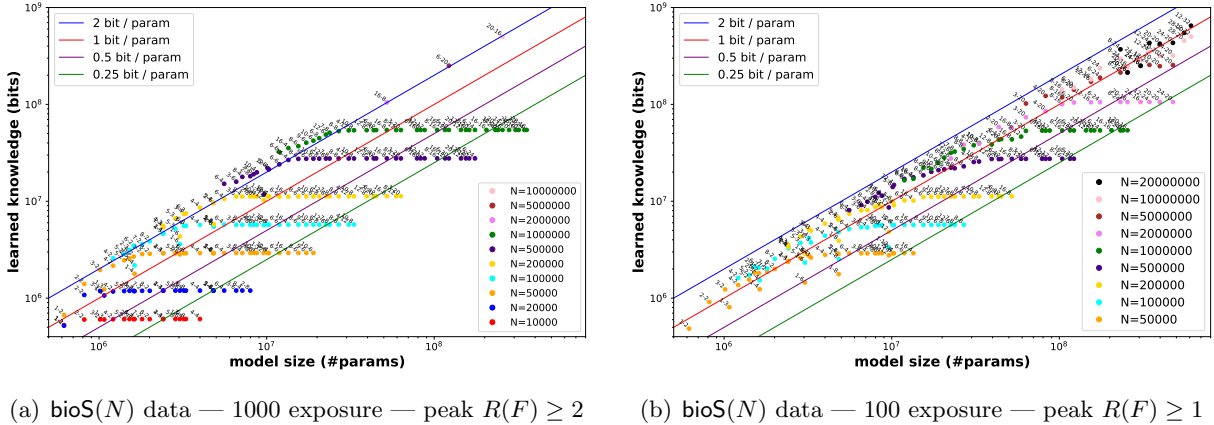

(a) bioS($N$) data — 1000 exposure — peak $R(F) \geq 2$  (b) bioS($N$) data — 100 exposure — peak $R(F) \geq 1$

Figure 9: Scaling laws for GPT2 pretrained on bioS($N$) data with fp16 (mixed-precision) for 1000/100 exposures, **now including 1-layer transformers** comparing to Figure 1. **Conclusion:** 1-layer transformers show a minor capacity ratio deficiency, especially in the 100-exposure setting.

parameters, our bioS($N$) data utilizes only 3275 tokens (after applying GPT2's tokenizer), reducing the effective embedding layer size to $3275 \times (64h)$. This adjustment explains why, for bioS data, GPT2small, typically known to have 124M parameters, is counted as having only 88M parameters in this paper.

We have selected a broad range of GPT2-$\ell$-$h$ models with practical $\ell$ and $h$ values, excluding those with similar model sizes. Their selection is detailed in Figure 1, encompassing both wide and shallow transformers (e.g., GPT2-2-20, GPT2-3-20, GPT2-4-20) and skinny and deep transformers (e.g., GPT2-16-4, GPT2-16-8, GPT2-28-20). For reference, GPT2 small/med/large correspond to GPT2-12-12, GPT2-24-16, GPT2-36-20, respectively.

We primarily focus on models with $\ell \geq 2$, as 1-layer transformers may demonstrate slightly lower capacity ratios. (For those interested, 1-layer transformers are included in Figure 9, which is identical to Figure 1 but includes these models.)

**Model sizes for datasets bioS($N$) with $N \geq 2M$.** In the 1000-exposure setting, to conserve computational resources, when exploring scaling laws for $N = 2M, 5M, 10M, 20M$, we concentrate on *one model size* per dataset — specifically GPT2-16-8, GPT2-6-20, GPT2-20-16, GPT2-25-20 — as they approach the 2bit/param threshold (i.e., they satisfy $R^{\mathsf{max}}(F) \approx 2$). In this context, our key finding is the validation of the 2bit/param capacity ratio, thus examining a limited selection of model sizes is adequate.

For the 100-exposure setting, we evaluate a broader range of model sizes per dataset. This approach is not only due to the tenfold reduction in training time compared to the 1000-exposure setting but also to facilitate a detailed comparison of model architectures in the 100-exposure setting, aiming for precision at higher model sizes.

**Training parameters.** We employ the AdamW optimizer with a cosine learning rate scheduler. This includes 1K steps of warmup, followed by a cosine decay of the learning rate from 1 to 0.1 times the reference rate. We use mixed-precision fp16 training unless otherwise stated.

## A.1 Base Scaling Laws

Our base scaling laws for the 1000-exposure and 100-exposure bioS($N$) data are presented in Figures 1(a) and 1(b), respectively.

For the 1000-exposure setting, the model's final performance is *not very sensitive* to learning rate choices due to sufficient training. The following parameters were chosen for generating Figure 1(a):

**Parameter 1** (Figure 1(a)). In the 1000-exposure setting for GPT2 models on bioS($N$) data:
- For $N = 10K$, we use $wd = 0.02$, $lr = 0.001$, and batch size 24 (about 140K training steps);
- For $N = 20K$, we use $wd = 0.02$, $lr = 0.001$, and batch size 48 (about 140K training steps);
- For $N = 50K$, we use $wd = 0.02$, $lr = 0.001$, and batch size 96 (about 175K training steps);
- For $N = 100K, 200K$, we use $wd = 0.02$, $lr = 0.001$, batch size 192 (about 175K, 349K training steps);
- For $N = 500K, 1M$, we use $wd = 0.01$, $lr = 0.0005$, batch size 192 (about 435K, 870K training steps);
- For $N = 2M$, we use $wd = 0.005$, $lr = 0.0003$, and batch size 1536 (about 220K training steps);
- For $N = 5M$, we use $wd = 0.002$, $lr = 0.0003$, and batch size 1536 (about 540K training steps);
- For $N = 10M$, we use $wd = 0.001$, $lr = 0.0003$, and batch size 1536 (about 1M training steps).

*Remark* A.1 (fp16 vs bf16). Training on GPT2 is conducted using mixed-precision fp16. We also tried bf16 and the results are nearly identical.

*Remark* A.2 (parameters). These optimization parameters are *very natural*, as it is generally impossible to have a fixed set of parameters for model sizes across a large multiplicative range. Notably:
- Larger model sizes naturally require smaller learning rates.

- Language models typically need at least 50K training steps *regardless of* batch size. Thus, for small $N$, we *reduce the batch size* to ensure the total number of training steps exceeds this threshold. For very large models, a larger batch size is preferred to enable GPU parallelism.

- When $lr$ remains constant, $wd$ should be relatively reduced as the number of training steps increases. Mathematically, the model weights should be "halved" for every $\Theta(\frac{1}{lr \times wd})$ training steps. Therefore, it's advisable to reduce the $wd$ parameter when training for longer periods.

*Remark* A.3 (# GPUs). In this paper, we do not specify the number of GPUs as it is irrelevant. The results remain the same whether using 64 GPUs each with a batch size of 24, 48 GPUs each with a batch size of 32, or 1536 GPUs each with a batch size of 1.

For the 100-exposure setting, careful tuning of learning rates is required. The following parameters were chosen for generating Figure 1(b): (Note: $N = 10K, 20K$ are not considered for the 100-exposure setting due to the excessively short training process.)

**Parameter 2** (Figure 1(b)). In the 100-exposure setting for GPT2 models on bioS($N$) data:
- For $N = 50K$, we use $wd = 0.01$, $lr = 0.001$, and batch size 12;
- For $N = 100K$, we use $wd = 0.01$, $lr = 0.001$, and batch size 24;
- For $N = 200K$, we use $wd = 0.01$, $lr = 0.001$, and batch size 48; (except for GPT2-2-20, where $lr = 0.0005$ is used)
- For $N = 500K$, we use $wd = 0.01$, $lr = 0.0005$, and batch size 96;
- For $N = 1M$, we use $wd = 0.01$, $lr = 0.0005$, and batch size 192;
- For $N = 2M$, we use $wd = 0.01$, $lr = 0.0003/0.0005/0.001$, and batch size 384;
- For $N = 5M$, we use $wd = 0.01$, $lr = 0.0003/0.0005$, and batch size 768;
- For $N = 10M$, we use $wd = 0.01$, $lr = 0.0002/0.0003/0.0005$, and batch size 1024;
- For $N = 20M$, we use $wd = 0.002$, $lr = 0.0002/0.0003/0.0005$, and batch size 1536.[22]

## A.2 Knowledge Memorization vs. Extraction

It was recently discovered by Allen-Zhu and Li [3] that although models memorize knowledge, this knowledge may not be extractable (e.g., via fine-tuning) for application in downstream tasks. It is essential to verify that the "2 bit/param" knowledge learned by models is indeed extractable. This verification is achieved by applying a fine-tuning task (e.g., "What is Anya's birthday? Answer: October 2, 1996") to half of the individuals and then testing its performance on the remainder.

Specifically, on the original bioS($N$) data, we compute two quantities for each model:

- MEMORIZABLE KNOWLEDGE ACCURACY (# OF PEOPLE).

  We apply the model to the original training data, such as "Anya Briar Forger was born on" and check if it can correctly generate "October 2, 1996". For each person, we evaluate all five attributes and compute their average accuracy.[23] We then <u>sum</u> this accuracy up over all $N$ people. (Ideally, a perfect model would have this "accuracy" equal to $N$.)

- EXTRACTABLE KNOWLEDGE ACCURACY (# OF PEOPLE).

  Following the pretrain-finetune framework of [3], we fine-tune any given pretrained model on half of the individuals using LoRA [17] with question-answering texts like "What is the birthday of Anya Briar Forger? Answer: October 2, 1996." We then test its generation accuracy on the remaining half of the individuals. High accuracy indicates that the knowledge is not only memorized but can also be *flexibly* extracted for downstream tasks. Again, for each person,

---
[22]Except for GPT2-28-20 we run out of GPU memory so reduce to batch size 1280.

[23]We exclude the company city attribute because it can be uniquely determined by the employer name, thus providing no additional knowledge.

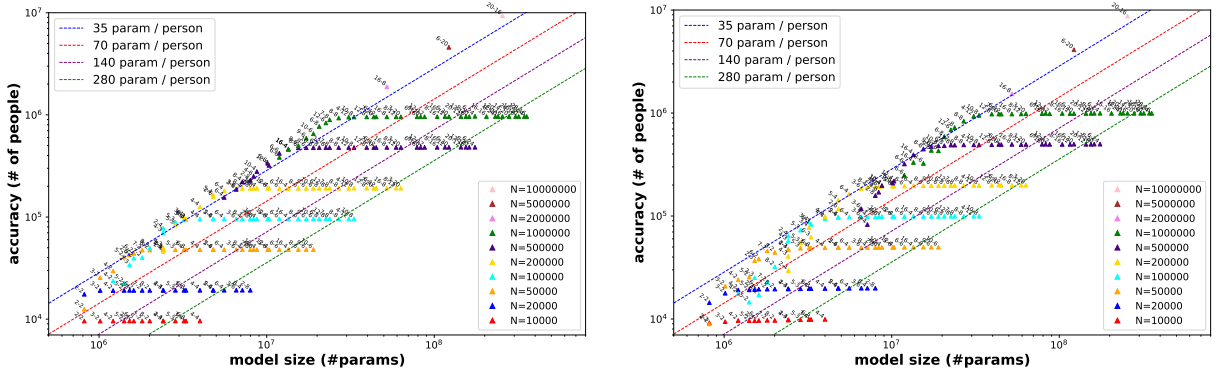

(a) **1000 exposures** — memorizable knowledge accuracy  (b) **1000 exposures** — extractable knowledge accuracy

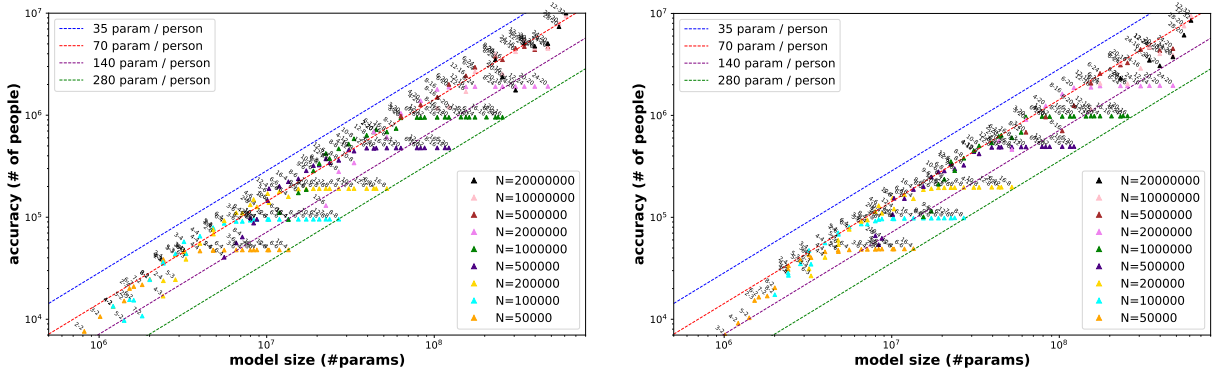

(c) **100 exposures** — memorizable knowledge accuracy  (d) **100 exposures** — extractable knowledge accuracy

Figure 10: Our scaling laws from Figure 1 also apply to *extractable knowledge* (see definitions in Section A.2). This figure is for the bioS($N$) datasets using GPT2 models.

*Remarks.* Recall from Remark 4.4 that each person's biography contains over 47.6 bits of knowledge (excluding names), explaining why the y-axis in this figure is $\sim 50$ times smaller than in Figure 1.

we evaluate all five attributes and compute their average accuracy. We then <u>sum</u> this across all $N/2$ people and multiply by 2. (Once again, a perfect model would have this equal to $N$.)

Our results are presented in Figure 10. By comparing, for instance, Figure 10(a) against Figure 10(b), it is evident that our scaling laws apply not only to memorizable knowledge but also largely to extractable knowledge. Only for models precisely at the capacity ratio boundary is there a 1.2x decrease in total accuracy.[24]

**Parameter 3** (Figure 10). When dealing with models of significantly different sizes for LoRA finetuning, it's necessary to adjust the LoRA rank sizes. In [3], the authors primarily used a rank $r' = 128$ update for the embedding layer and ranks $r = 8$ or 16 for the query/value matrices, with their base model being either GPT2-12-12 or GPT2-12-20. In this paper, we explore a broader range of rank choices: $(r', r) \in \{(8, 2), (16, 2), (8, 4), (32, 4), (8, 8), (32, 8), (128, 8), (32, 16), (128, 16)\}$, presenting only the best results.[25]

---

[24]This decrease is in accuracy, not bits; a model may have a large amount of extractable knowledge in bits but not in accuracy. One can also compute knowledge *bits* in the extractable setting, but we omit such results for brevity.

[25]Selecting the best LoRA option is justified as our aim is to determine the maximum extractable knowledge bits, and thus, any LoRA option demonstrating high test-set accuracy fulfills our objective.

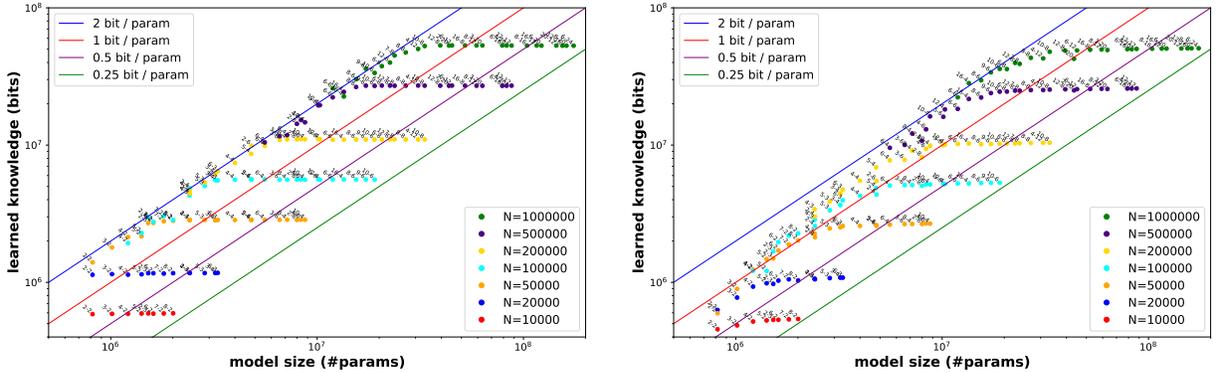

(a) on bioS$^{\text{simple}}$ data that has no sentence diversity  (b) on the semi-real bioR data generated by LLaMA2

Figure 11: Scaling laws for the bioS$^{\text{simple}}$ and bioR data with **1000 exposures**.

We disable learning rate warmup, set the batch size to 96, the learning rate to 0.001 (with linear decay down to 0), weight decay at 0.1, and finetune for 75,000 steps.

### A.3 Other Biography Datasets

We also examine the bioS$^{\text{simple}}$($N$) datasets, which are identical to bioS($N$) except that each individual's knowledge is stored in a fixed ordering of six fixed sentences (see Section 2.2). Allen-Zhu and Li [3] found that in such cases, the knowledge data are memorizable but nearly 0% extractable. As shown in Figure 11(a), in these instances, the capacity ratio slightly decreases compared to Figure 1(a). This implies, in this ideal setting, adding data diversity — by rewriting the same knowledge multiple times using different writing templates — not only enhances the model's ability to extract knowledge, as noted by [3], but also, surprisingly, *increases* the model's capacity, as observed in this study.

Moreover, we explore the semi-real dataset bioR($N$), which resembles bioS($N$) but with the biography paragraph generated by LLaMA2, and each individual is generated 40 times (using random seeds and prompts to encourage LLaMA2 to generate as diverse paragraphs as possible for each person). This results in a total of 22GB of text, comparable to the size of Wikipedia data.

The scaling law for the bioR($N$) data is presented in Figure 11(b), indicating that the capacity ratio slightly decreases for larger models. This trend is expected, as LLaMA2 introduces numerous irrelevant details into the human biographies — usually different irrelevant details for each LLaMA2 generation — thereby consuming more model capacity. The decrease is more significant for smaller models, which may have greater difficulty comprehending the diverse English sentences in the data.

**Parameter 4** (Figure 11). In both experiments, we adhere to the same set of optimizer parameters used in Figure 1(a), as detailed in Parameter 1.

### A.4 More on Parameterized Scaling Laws

In the parameterized scaling laws, we utilize the bioD($N, K, C, D, L, T$) dataset from Definition 2.2.

**Parameter 5** (Figure 2, 12, 13). For GPT2 models on the bioD dataset, we focus on the 1000-exposure case, with $wd = 0.01$, $lr = 0.0005$, and a batch size of 192.

*Remark* A.4 (parameters). Contrary to Parameter 1, it is not necessary to vary the training parameters, as our experiments with GPT2 models span a much narrower range of model sizes. We have adjusted the choice of $N$ to ensure that the optimal 2bit/param models are within a factor of 20 of each other in terms of model sizes.

Our results are presented in Figure 2 (in the main body, limited to models with accuracy $\leq 50\%$ for clarity) and in Figure 12 (including all models).

Furthermore, from the bit complexity lower bound (see Definition 4.1)

$$\underbrace{N \log_2 \frac{N_0}{e^{p_1}}}_{\text{name}} + \underbrace{NK \log_2 \frac{D^C}{e^{p_2}}}_{\text{value}} + \underbrace{KD \log_2 \frac{T^L}{De^{p_3}}}_{\text{diversity}} \tag{A.1}$$

we also dissect how the three components contribute to this overall lower bound. As shown in Figure 13, although the "value" component typically dominates, for certain hyperparameter settings, the "name" or "diversity" components can also be significant. This underscores the importance of proving our Theorem 3.2 lower bound, which is a sum of all three terms.

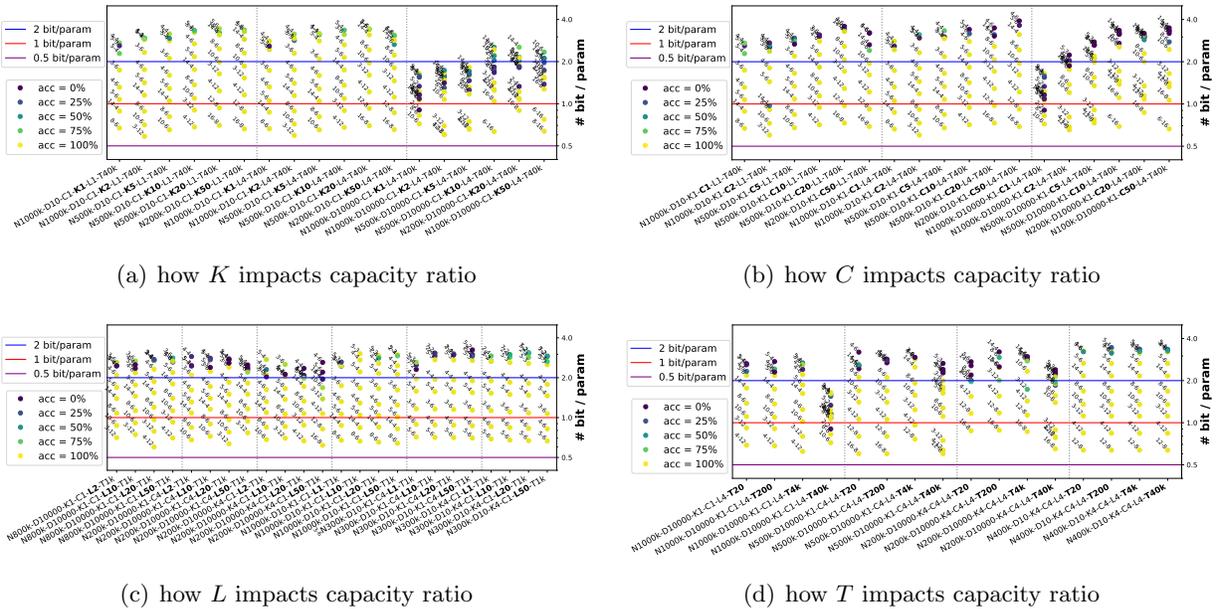

(a) how $K$ impacts capacity ratio

(b) how $C$ impacts capacity ratio

(c) how $L$ impacts capacity ratio

(d) how $T$ impacts capacity ratio

Figure 12: Same as Figure 2, but including models with accuracies below 50% (which may overlap with higher-accuracy models). The *peak* capacity ratios consistently exceed $R(F) \geq 2$.

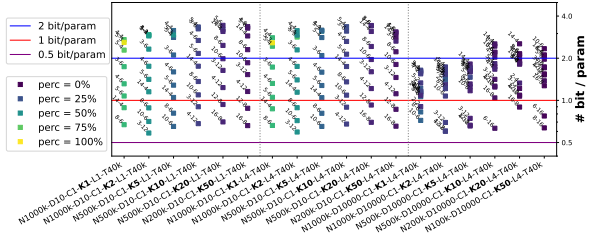
(a) percentage of bits belonging to "name" in (A.1)

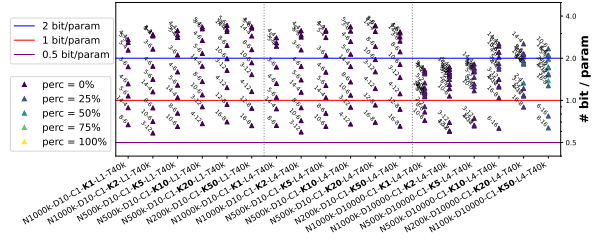
(b) percentage of bits belonging to "diversity" in (A.1)

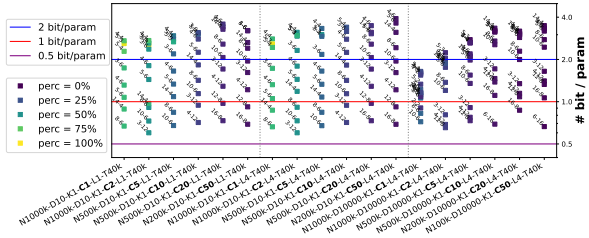
(c) percentage of bits belonging to "name" in (A.1)

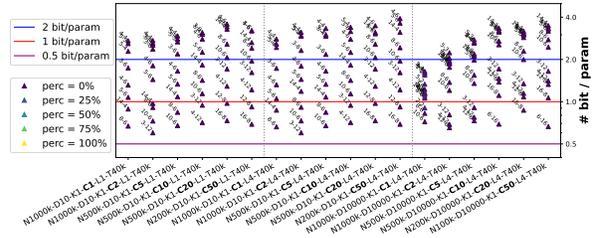
(d) percentage of bits belonging to "diversity" in (A.1)

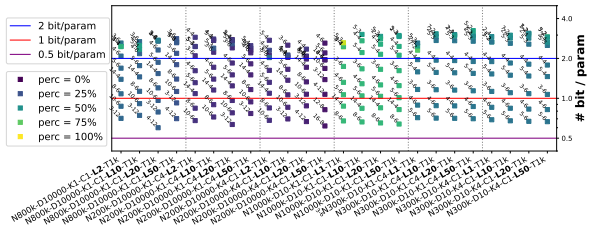
(e) percentage of bits belonging to "name" in (A.1)

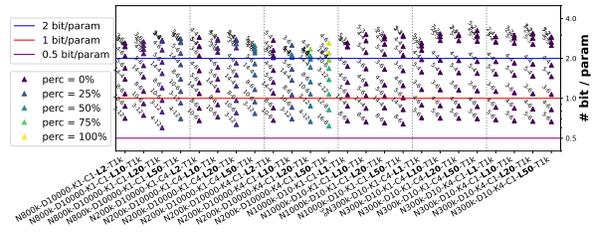
(f) percentage of bits belonging to "diversity" in (A.1)

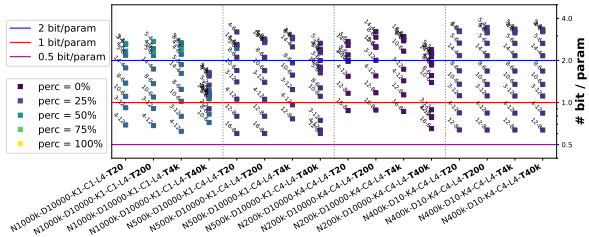
(g) percentage of bits belonging to "name" in (A.1)

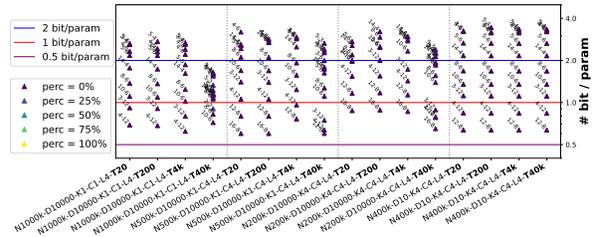
(h) percentage of bits belonging to "diversity" in (A.1)

Figure 13: Breakdown of knowledge components in the parameterized bioD scaling law experiments, as shown in Figure 2. Refer to Equation (A.1) and the accompanying text.

# B  More on Model Architectures

We explore alternative architectural choices for language models.

**LLaMA/Mistral.** Notably, as of the writing of this paper, LLaMA [32, 33] and Mistral [19] stand out as popular, publicly-available large language models. We highlight their key architecture differences from GPT2 — which we define as having rotary embedding and no dropout.

1. LLaMA and Mistral employ MLP layers with gated activation, using $V(\sigma(W_1 x) \cdot (W_2 x))$ instead of $V\sigma(Wx)$. Shazeer [29] noted that gated activation appears to yield slightly better performance.

2. Unlike GPT2, which ties the weights of the embedding layer and the output (LMHead) layer, LLaMA and Mistral do not.

3. For a hidden dimension $d$, GPT2/LLaMA have $4d^2$ parameters in the attention layer and $8d^2$ in the MLP layer, whereas Mistral allocates a larger $10.5d^2$ for its MLP layer.

4. Mistral promotes group-query attention (e.g., using 4 groups, thus reducing the K/V matrices to $d^2/4$ in size), unlike GPT2. LLaMA does not favor multi-query attention unless in its very large models, such as the 70B variant.

5. LLaMA and Mistral utilize different tokenizers compared to GPT2, with Mistral's tokenizer being nearly identical to LLaMA's.

6. GPT2 employs $\sigma = gelu$, while LLaMA/Mistral use $\sigma = silu$.

7. GPT2 incorporates layer normalization with trainable bias, which LLaMA/Mistral do not.

Given these distinctions, for LLaMA models, we use the notation LLaMA-$\ell$-$h$ for $\ell$ layers, $h$ heads, and $64h$ hidden dimensions; we omit group-query attention as LLaMA recommends it only for its 70B model. For Mistral, denoted as Mistral-$\ell$-$h$, we enable group-query attention with 4 groups if $h = 0 \pmod 4$, 1 group for odd $h$, or 2 groups otherwise.

**GPT2 with Smaller MLP.** Mistral has a larger MLP layer, and it is often believed that the MLP layer serves primarily for storing knowledge, in contrast to the Attention layer. But is this truly the case?

To delve into this, we examine $\text{GPT2}_{1/4}$, which is GPT2 with its MLP layer reduced from $d \to 4d \to d$ to $d \to d \to d$ (thus, 1/4 of its original size), and $\text{GPT2}_0$, which is GPT2 but without any MLP layer.

**Experimental setups.** Throughout this section, when presenting positive result (such as for GPT2) we try to stick to one fixed set of learning rate choices; but when presenting a negative result (such as for the LLaMA architecture), we present the best among three learning rate choices.

## B.1  1000-Exposure Setting

In the 1000-exposure setting, we observe that the model architecture choices have a *negligible impact* on the scaling laws. The results for LLaMA, Mistral, $\text{GPT2}_0$, and $\text{GPT2}_{1/4}$ architectures are presented in Figure 3, with their parameter choices discussed below.

**Parameter 6** (Figure 3)**.** In the 1000-exposure setting, for LLaMA/Mistral models we use similar parameters as specified in Parameter 1, but we select the best of three learning rates to better demonstrate that GPT2 performs *no worse* than *even the best tuned* LLaMA/Mistral models:

- For $N = 10K$, we use $wd = 0.02$, $lr = 0.0005/0.001/0.002$, and batch size 24 with fp16;
- For $N = 20K$, we use $wd = 0.02$, $lr = 0.0005/0.001/0.002$, and batch size 48 with fp16;

- For $N = 50K$, we use $wd = 0.02$, $lr = 0.0005/0.001/0.002$, and batch size 96 with fp16;
- For $N = 100K, 200K$, we use $wd = 0.02$, $lr = 0.0005/0.001/0.002$, and batch size 192 with fp16;
- For $N = 500K, 1M$, we use $wd = 0.01$, $lr = 0.0002/0.0003/0.0005$, and batch size 192 with fp16;
- For $N = 2M$, we use $wd = 0.005$, $lr = 0.0003/0.0005/0.001$, and batch size 1536 with bf16;
- For $N = 5M$, we use $wd = 0.002$, $lr = 0.0003/0.0005/0.001$, and batch size 1536 with bf16;
- For $N = 10M$, we use $wd = 0.001$, $lr = 0.0003/0.0005/0.001$, and batch size 1536 with bf16.

For $GPT2_0$ and $GPT2_{1/4}$, we use the same learning rates as specified in Parameter 1.

*Remark* B.1 (bf16 on gated MLP). As discussed in Section B.2, the training of LLaMA and Mistral architectures is less stable due to the use of GatedMLP, leading to the necessity of switching to (mixed-precision) bf16 training when required.

From Figure 3, it is evident that, except for tiny models, LLaMA, Mistral, $GPT2_0$, and $GPT2_{1/4}$ architectures closely follow GPT2's scaling law over 1000 exposures. For tiny models with $\leq 10M$ parameters, tying model weights increases their capacity (refer to Figure 3(c)). This indicates that the *2bit/param capacity ratio is a relatively universal law* among most typical (decoder-only) language model architectures.

## B.2  100-Exposure Setting

The 100-exposure setting reveals more intriguing comparisons. We contrast GPT2 with various model architectures in Figure 4 and offer a detailed comparison between LLaMA and GPT2 architectures in Figure 5.

Figure 4(b) shows that the LLaMA architecture may lag behind GPT2's scaling law by a factor of 1.3x, even for larger models.

We delve into the reasons behind this. By adjusting LLaMA's architecture (e.g., switching GatedMLP back to normal MLP), as shown in Figure 5, we find that replacing LLaMA's GatedMLP with a standard MLP is necessary to match GPT2's scaling law. Notably, for a strong comparison, when using GatedMLP we select the best result from three learning rates, whereas for a standard MLP, akin to GPT2, we use a single learning rate. For smaller models, matching GPT2 requires tying model weights and adopting GPT2's tokenizer, though this is less significant.[26]

For other model architectures, Mistral, $GPT2_0$, and $GPT2_{1/4}$, their scaling laws in the 100-exposure setting are presented in Figure 4. Figure 4(c) confirms that the Mistral architecture also underperforms GPT2 due to its use of gated MLP. Figure 4(d) reveals that reducing $GPT2_{1/4}$'s MLP layer size by a quarter has a *negligible impact* on model capacity. However, removing the MLP layers entirely in $GPT2_0$ significantly reduces the model's capacity, see Figure 4(e).

The 100-exposure setting represents an "insufficient training" paradigm. Thus, the comparisons are not about one architecture being strictly worse than another (as they achieve similar capacity ratios in a 1000-exposure setting, as shown in Figure 3). Our findings indicate that some architectures are *noticeably easier to train (thus learn knowledge faster)*:

- The GatedMLP architecture *slows down* the model's learning speed, and we observe less stable training with its use.[27]
- Removing MLP layers entirely *slows down* the model's learning speed, whereas adjusting the size of MLP layers (e.g., from $8d^2$ to $10.5d^2$ or down to $2d^2$) may not have a significant impact.

---

[26]The influence of the tokenizer on model capacity is noteworthy. For instance, LLaMA/Mistral tokenizers tend to split birthday years into single-digit tokens, slightly slowing the training of smaller models, whereas the GPT2Tokenizer uses a single token for the birth years such as 1991.

[27]For example, mixed-precision fp16 training can sometimes fail for LLaMA/Mistral models smaller than 100M; hence, we use mixed-precision bf16 instead. Conversely, GPT2 models up to 1B can be trained with fp16.

Additionally, we experimented with enabling trainable biases in LLaMA's layernorms and switching from *silu* to *gelu* (to more closely resemble GPT2), in a similar way as Figure 5, but found these changes do not affect the model's capacities. We ignore those experiments for clarity.

Below, we discuss our parameter choices for the experiments in Figure 4 and Figure 5.

**Parameter 7** (Figure 4). In the 100-exposure setting,

(a) For LLaMA/Mistral models on bioS($N$) data, aiming to present *negative* results, we select the best learning rate from three options in each data setting:
- For $N = 50K$, we use $wd = 0.01$, $lr = 0.0003/0.0005/0.001$, and batch size 12 with bf16;
- For $N = 100K$, we use $wd = 0.01$, $lr = 0.0003/0.0005/0.001$, and batch size 24 with bf16;
- For $N = 200K$, we use $wd = 0.01$, $lr = 0.0003/0.0005/0.001$, and batch size 48 with bf16;
- For $N = 500K$, we use $wd = 0.01$, $lr = 0.0002/0.0003/0.0005$, and batch size 96 with bf16;
- For $N = 1M$, we use $wd = 0.01$, $lr = 0.0002/0.0003/0.0005$, and batch size 192 with bf16;
- For $N = 2M$, we use $wd = 0.01$, $lr = 0.0003/0.0005/0.001$, and batch size 384 with bf16;
- For $N = 5M$, we use $wd = 0.01$, $lr = 0.0003/0.0005/0.001$, and batch size 768 with bf16;
- For $N = 10M$, we use $wd = 0.01$, $lr = 0.0003/0.0005/0.001$, and batch size 1536 with bf16;
- For $N = 20M$, we use $wd = 0.002$, $lr = 0.0003/0.0005/0.001$, and batch size 1536 with bf16.

(For $N \leq 1M$, we also tested the same settings with fp16, finding similar results. However, LLaMA/Mistral models tend to fail more often with fp16, so we primarily used bf16.)

(b) For GPT2$_{1/4}$:
- For $N = 50K$, we use $wd = 0.01$, $lr = 0.0005/0.001$, and batch size 12 with fp16;
- For $N = 100K$, we use $wd = 0.01$, $lr = 0.0005/0.001$, and batch size 24 with fp16;
- For $N = 200K$, we use $wd = 0.01$, $lr = 0.0005/0.001$, and batch size 48 with fp16;
- For $N = 500K$, we use $wd = 0.01$, $lr = 0.0003/0.0005$, and batch size 96 with fp16;
- For $N = 1M$, we use $wd = 0.01$, $lr = 0.0003/0.0005$, and batch size 192 with fp16.

(c) For GPT2$_0$, to present a *negative* result, we use the same settings as in Parameter 4(a):
- For $N = 50K$, we use $wd = 0.01$, $lr = 0.0003/0.0005/0.001$, and batch size 12 with bf16;
- For $N = 100K$, we use $wd = 0.01$, $lr = 0.0003/0.0005/0.001$, and batch size 24 with bf16;
- For $N = 200K$, we use $wd = 0.01$, $lr = 0.0003/0.0005/0.001$, and batch size 48 with bf16;
- For $N = 500K$, we use $wd = 0.01$, $lr = 0.0002/0.0003/0.0005$, and batch size 96 with bf16;
- For $N = 1M$, we use $wd = 0.01$, $lr = 0.0002/0.0003/0.0005$, and batch size 192 with bf16.

**Parameter 8** (Figure 5). In the 100-exposure controlled comparison experiment,
- For presenting *negative* results (Figure 5(a) and Figure 5(c)), we select the best learning rate from three options, identical to GPT2$_0$ in Parameter 4(c).
- For presenting *positive* results (Figure 5(b) and Figure 5(d)), we use a single set of learning rates, identical to Parameter 2 but with fp16 replaced by bf16 for a stronger comparison.

# C  More on Quantization

We use the `auto_gptq` package (based on [10]) to quantize the GPT2 model results in Figure 1 for the bioS data and the GPT2 model results in Figure 2 for the bioD data. We simply use a small set of 1000 people's biographies to perform the quantization task. Our results are presented in Figure 14 for the bioS data and in Figure 15 for the bioD data.

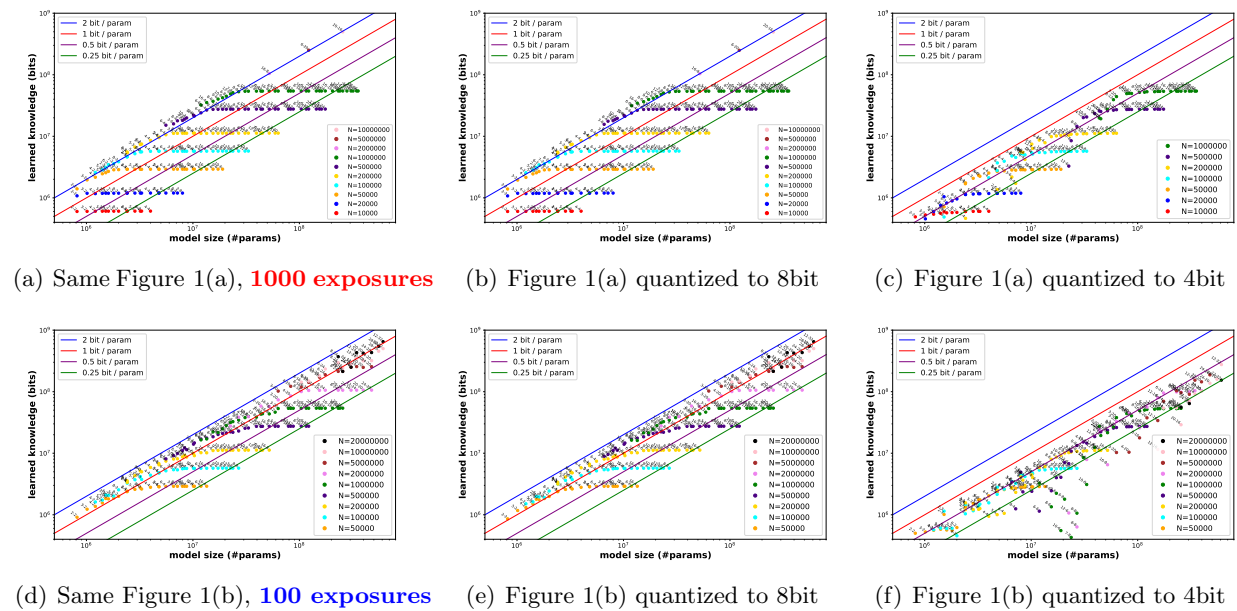

(a) Same Figure 1(a), **1000 exposures**   (b) Figure 1(a) quantized to 8bit   (c) Figure 1(a) quantized to 4bit

(d) Same Figure 1(b), **100 exposures**   (e) Figure 1(b) quantized to 8bit   (f) Figure 1(b) quantized to 4bit

Figure 14: Scaling laws for GPT2 after quantizing Figure 1 into int8 and int4.

> **Conclusion.** Quantizing a mixed-precision fp16 trained model into int8 shows no change, but quantizing into int4 results in a capacity ratio loss greater than 2x.

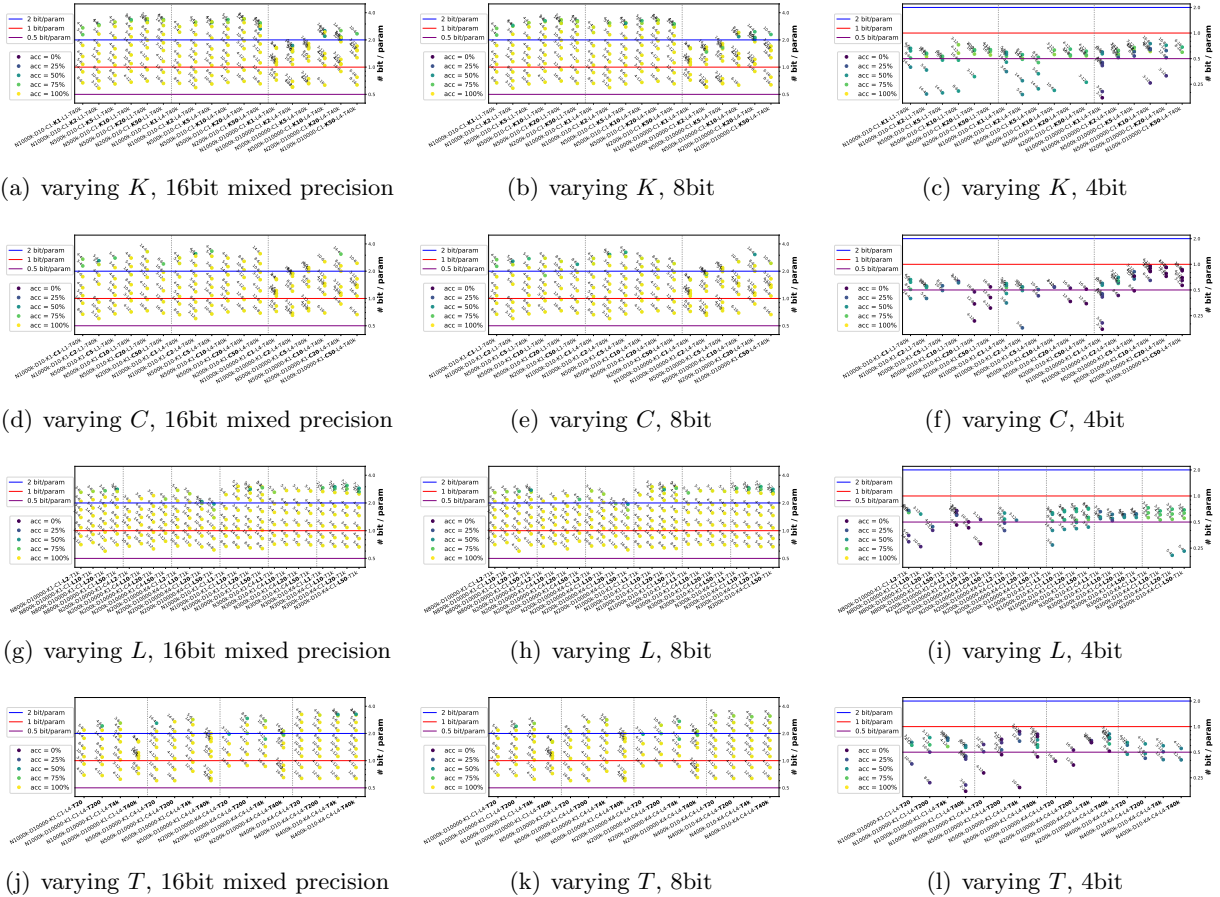

Figure 15: 8-bit/4-bit quantization of GPT2 models trained on bioD($N, K, C, D, L, T$) data for **1000 exposures**. **Left:** Identical to Figure 2, showing only models with accuracy $\geq 50\%$; **Middle:** After quantization to 8-bit; **Right:** After quantization to 4-bit, including models with all accuracies.

**Observation:** For the bioD data family, quantizing to 8-bit has negligible impact on model capacities. Quantizing to 4-bit reduces capacity by more than 2x, especially for large $D$ and $L$, leading to significantly larger reductions and explaining the missing columns in Figure 15(i)).

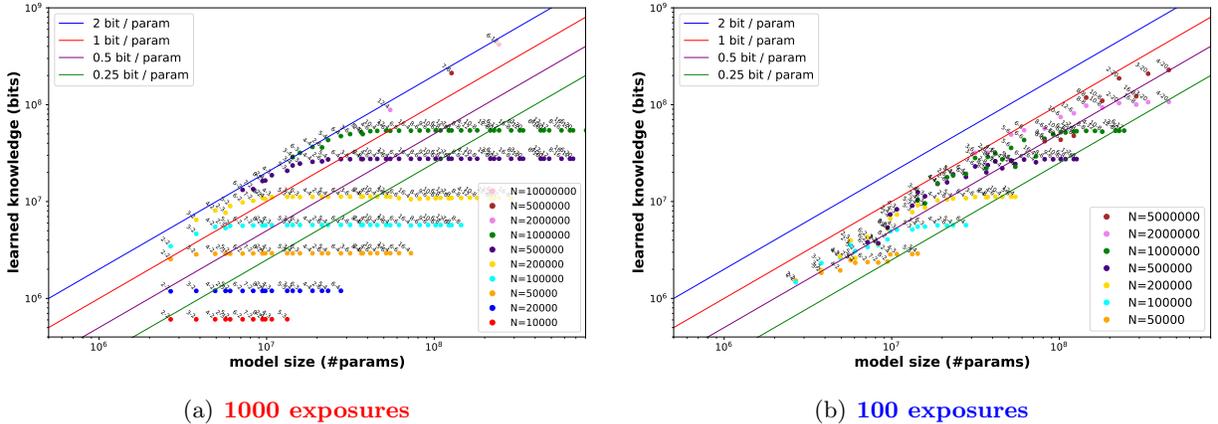

(a) **1000 exposures**  (b) **100 exposures**

Figure 16: Scaling laws for the mixture-of-experts GPT2 models with 32 experts on the bioS($N$) data.

> **Remarks.** Compared to Figure 1, using MoE with 32 experts reduces the 1000-exposure capacity by 1.3x and the 100-exposure one by 1.5x, despite the MoE model using 11.3x fewer parameters during inference. For the strongest result we use $topk = 1, cap\_factor = 2$; other variants are in Figure 17.

## D  More on Mixture of Experts

We utilize the `tutel` package for implementing Mixture-of-Experts (MoE) on GPT2 models [18]. In MoE, the parameter *topk* determines the number of experts each token is routed to. It is recommended by some practitioners to use $topk = 2$ during training and $topk = 1$ during testing. Additionally, the $cap\_factor$ parameter ensures that, given $M$ experts, each expert receives no more than $\frac{cap\_factor}{M}$ fraction of the data.

Using $topk = 1$ and $cap\_factor = 1$ is generally not advisable. Thus, to provide the strongest result, we set $topk = 1, cap\_factor = 2$ for the 1000/100-exposure scaling laws in Figure 16. (During testing, we increase the capacity factor to $cap\_factor = 8$.)

For the 100-exposure scaling law, we additionally compare three configurations: $(topk, cap\_factor) = (1, 2), (2, 1), (2, 2)$, finding minimal differences among them as shown in Figure 17. Remember from Section 7 that differences in model architecture usually become apparent in the insufficient training regime; this is why we opt for 100-exposure instead of 1000-exposure. Notably, $(topk, cap\_factor) = (2, 2)$ performs best (among the three) for deep models, such as GPT2-16-4 with 32 experts.

Due to their sparsity, MoE models often require higher learning rates compared to dense models. Consequently, we adjust the optimizer parameters as follows:

**Parameter 9** (Figure 16, Figure 17)**.** In the 1000-exposure setting for GPT2-MoE models with 32 experts, we slightly increase the learning rates while keeping other parameters nearly identical to Parameter 1:
- For $N = 10K$, we use $wd = 0.02$, $lr = 0.001/0.002$, and batch size 24 with fp16;
- For $N = 20K$, we use $wd = 0.02$, $lr = 0.001/0.002$, and batch size 48 with fp16;
- For $N = 50K$, we use $wd = 0.02$, $lr = 0.001/0.002$, and batch size 96 with fp16;
- For $N = 100K, 200K$, we use $wd = 0.02$, $lr = 0.001/0.002$, batch size 192 with fp16;
- For $N = 500K, 1M$, we use $wd = 0.01$, $lr = 0.0005/0.001$, batch size 192 with fp16;
- For $N = 2M$, we use $wd = 0.005$, $lr = 0.002$, and batch size 1536 with fp16;
- For $N = 5M$, we use $wd = 0.002$, $lr = 0.0005$, and batch size 1536 with fp16;
- For $N = 10M$, we use $wd = 0.001$, $lr = 0.0005$, and batch size 1536 with fp16.

In the 100-exposure setting, we also use higher learning rates compared to Parameter 2:
- For $N = 50K$, we use $wd = 0.01$, $lr = 0.001/0.002/0.005$, and batch size 12 with fp16;

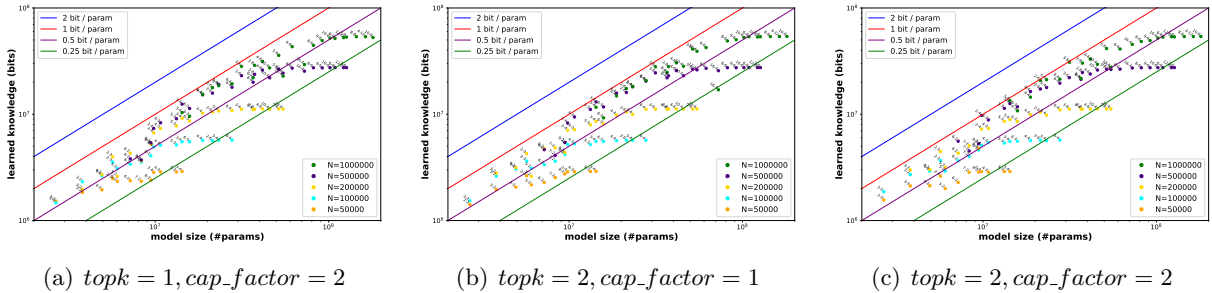

|   (a) $topk = 1, cap\_factor = 2$   |   (b) $topk = 2, cap\_factor = 1$   |   (c) $topk = 2, cap\_factor = 2$   |

Figure 17: Scaling laws for the GPT2 MoE models with 32 experts on the $\mathsf{bioS}(N)$ data for **100 exposures**. This figure complements Figure 16 by comparing the effects of varying $topk$ and $cap\_factor$ in the 100-exposure insufficiently-trained regime. **Conclusion:** minimal differences are observed across these settings, though deeper models (e.g., GPT2-16-4 with 32 experts) seem easier to train with $topk = cap\_factor = 2$.

- For $N = 100K$, we use $wd = 0.01$, $lr = 0.001/0.002/0.005$, and batch size 24 with fp16;
- For $N = 200K$, we use $wd = 0.01$, $lr = 0.001/0.002/0.005$, and batch size 48 with fp16;
- For $N = 500K$, we use $wd = 0.01$, $lr = 0.001/0.002$, and batch size 96 with fp16;
- For $N = 1M$, we use $wd = 0.01$, $lr = 0.0005/0.001/0.002$, and batch size 192 with fp16;
- For $N = 2M$, we use $wd = 0.005$, $lr = 0.0005/0.001$, and batch size 192 with fp16;
- For $N = 5M$, we use $wd = 0.005$, $lr = 0.0003/0.0005/0.001$, and batch size 384 with fp16.

## E  More on Junk Data vs. Scaling Laws

Recall from Section 10 that our dataset is a mixture, with $1/8$ of the tokens coming from $\mathsf{bioS}(N)$ for various $N$ (referred to as "useful data"), and the remaining $7/8$ from "junk data." We explored three scenarios:

(a) Junk data being $\mathsf{bioS}(N')$ for $N' = 100M$, representing completely random junk;

(b) Junk data being $\mathsf{bioS}(N')$ for $N' = 1K$, representing highly repetitive data; and

(c) Junk data being $\mathsf{bioS}(N')$ for $N' = 100M$, but with a special token appended to the front of each piece of *useful data*.[28]

For simplicity, within each 512-token context window, we either include only useful data or only junk data (separated by `<EOS>` tokens). The outcomes are similar when mixing useful and junk data in the same context window. In all three cases, we initially consider a 100-exposure training setting where the useful data receive 100 exposures each during pretraining — thus, the total number of training tokens is approximately 8 times more than in Figure 1(b) (our scaling law for the 100-exposure case without junk data).

In case (A), presenting a *negative result*, we also explore 300-exposure, 600-exposure, and 1000-exposure training settings. Given that the 1000-exposure setting requires 48x more training tokens compared to Figure 1(b), or 4.8x more compared to Figure 1(a), we limited experiments to $\mathsf{bioS}(N)$ with $N \leq 200K$ to conserve computational resources. Similarly, for 300-exposure and 600-exposure, we only considered $N \leq 500K$.

In case (B), presenting a *positive result*, we limited our consideration to 100-exposure with $N \leq 1M$.

---

[28]This is akin to adding a domain name like wikipedia.org at the beginning of the data; the model lacks prior knowledge that these special token data signify high-quality, useful data. It's up to the model and the training process to *autonomously* discover this.

In case (C), presenting a *moderately positive result*, we explored both 100-exposure and 300-exposure settings, where, in the 300-exposure setting, we again limited to $N \leq 500K$.

Overall, due to the significantly different training durations (i.e., number of training tokens) across the 100-, 300-, 600-, and 1000-exposure settings, we had to adjust their batch sizes, weight decay, and learning rates accordingly. These adjustments are discussed below.

**Parameter 10** (Figure 8). We adhere to the general advice provided in Remark A.2 for selecting parameters in all experiments shown in Figure 8. For negative results (e.g., Figure 8(b), 8(c)), we opted for a smaller batch size to increase the number of trainable steps and explored a wider range of learning rate options. Conversely, for positive results (e.g., Figure 8(f), 8(e)), we sometimes chose a larger batch size to benefit from faster, GPU-accelerated training times and considered a narrower set of learning rate choices. Overall, we have meticulously selected parameters to strengthen negative results as much as possible while intentionally not optimizing positive results to the same extent. This approach ensures a *stronger comparison* and effectively communicates the key message of this section. **Specifically,**

- For Figure 8(b) which is Case (a) of 100-exposure:
  - For $N = 50K$, we use $wd = 0.01$, $lr = 0.0003/0.0005/0.001$, and batch size 12;
  - For $N = 100K$, we use $wd = 0.01$, $lr = 0.0003/0.0005/0.001$, and batch size 24;
  - For $N = 200K$, we use $wd = 0.01$, $lr = 0.0003/0.0005/0.001$, and batch size 48;
  - For $N = 500K$, we use $wd = 0.005$, $lr = 0.00005/0.0001/0.0002/0.0003/0.0005$, and batch size 192;
  - For $N = 1M$, we use $wd = 0.005$, $lr = 0.00005/0.0001/0.0002/0.0003/0.0005$, and batch size 192.
- For Figure 8(c) which is Case (a) of 300-exposure:
  - For $N = 50K$, we use $wd = 0.01$, $lr = 0.0003/0.0005/0.001$, and batch size 96;
  - For $N = 100K$, we use $wd = 0.01$, $lr = 0.0003/0.0005/0.001$, and batch size 192;
  - For $N = 200K$, we use $wd = 0.01$, $lr = 0.0003/0.0005/0.001$, and batch size 192;
  - For $N = 500K$, we use $wd = 0.01$, $lr = 0.0003/0.0005/0.001$, and batch size 192.
- For Figure 8(d) which is Case (a) of 600-exposure:
  - For $N = 50K$, we use $wd = 0.01$, $lr = 0.0003/0.0005/0.001$, and batch size 384;
  - For $N = 100K$, we use $wd = 0.01$, $lr = 0.0003/0.0005/0.001$, and batch size 384;
  - For $N = 200K$, we use $wd = 0.01$, $lr = 0.0003/0.0005/0.001$, and batch size 384;
  - For $N = 500K$, we use $wd = 0.002$, $lr = 0.0003/0.0005/0.001$, and batch size 768.
- For Figure 8(e) which is Case (a) of 1000-exposure:
  - For $N = 50K$, we use $wd = 0.01$, $lr = 0.0005/0.001$, and batch size 384;
  - For $N = 100K$, we use $wd = 0.01$, $lr = 0.0005/0.001$, and batch size 768;
  - For $N = 200K$, we use $wd = 0.01$, $lr = 0.0005/0.001$, and batch size 1536.
- For Figure 8(f) which is Case (b) of 100-exposure:
  - For $N = 50K$, we use $wd = 0.01$, $lr = 0.0003/0.0005$, and batch size 12;
  - For $N = 100K$, we use $wd = 0.01$, $lr = 0.0003/0.0005$, and batch size 24;
  - For $N = 200K$, we use $wd = 0.01$, $lr = 0.0003/0.0005/0.001$, and batch size 96;
  - For $N = 500K$, we use $wd = 0.01$, $lr = 0.0003/0.0005$, and batch size 192;
  - For $N = 1M$, we use $wd = 0.01$, $lr = 0.0003$, and batch size 192.
- For Figure 8(g) which is Case (c) of 100-exposure:
  - For $N = 50K$, we use $wd = 0.01$, $lr = 0.0003/0.0005/0.001$, and batch size 12;
  - For $N = 100K$, we use $wd = 0.01$, $lr = 0.0003/0.0005/0.001$, and batch size 24;
  - For $N = 200K$, we use $wd = 0.01$, $lr = 0.0002/0.0003/0.0005/0.001$, and batch size 96;
  - For $N = 500K$, we use $wd = 0.005$, $lr = 0.0002/0.0003/0.0005$, and batch size 192;
  - For $N = 1M$, we use $wd = 0.005$, $lr = 0.0002/0.0003/0.0005$, and batch size 192.
- For Figure 8(h) which is Case (c) of 300-exposure:
  - For $N = 50K$, we use $wd = 0.01$, $lr = 0.0003/0.0005/0.001$, and batch size 96;
  - For $N = 100K$, we use $wd = 0.01$, $lr = 0.0003/0.0005/0.001$, and batch size 192;

- For $N = 200K$, we use $wd = 0.01$, $lr = 0.0003/0.0005/0.001$, and batch size 192;
- For $N = 500K$, we use $wd = 0.005$, $lr = 0.0003/0.0005/0.001$, and batch size 384.

# F  Proof of Theorem 3.2

We present a crucial lemma that establishes the bit complexity required to encode random variables based on the probability that these variables match specific reference values.

**Lemma F.1.** *Let $\mathcal{Q}_1, \ldots, \mathcal{Q}_k$ be fixed sets (we call **domains**), and assume that for each $i \in [k]$, $Q_i$ is independently and randomly chosen from its corresponding domain $\mathcal{Q}_i$. Denote $Q = (Q_1, \ldots, Q_k)$ and view $Q$ as the **training data**.*

*Assume there exists a function $W(Q) \in \mathcal{W}$, which we regard as the **parameters** of a model computed (i.e., trained) from the training data $Q$.*

*Furthermore, consider an **evaluation function** $F_i$ that predicts*

$$\forall i \in [k]: \quad P_i = F_i(W(Q), Q_1, Q_2, \cdots, Q_{i-1}, R) \quad \text{with} \quad p_i(Q) \stackrel{\text{def}}{=} \Pr_R[P_i = Q_i \mid Q] \ .$$

*Here, $F$ is parameterized by $W(Q)$ and may rely on previous data $Q_1, \ldots, Q_{i-1}$, and new randomness $R$. Then, it follows that*

$$\log |\mathcal{W}| \geq \sum_{i \in [k]} \log \left( \mathbb{E}_Q[p_i(Q)] \times |\mathcal{Q}_i| \right) \geq \mathbb{E}_Q \left[ \sum_{i \in [k]} \log \left( p_i(Q) \times |\mathcal{Q}_i| \right) \right] \ . \tag{F.1}$$

*Proof of Lemma F.1.* Since the second inequality of (F.1) trivially comes from Jensen's inequality, we only prove the first one.

When $i = 1$, we have $P_1 = F_1(W(Q), R)$ and one can prove the lemma by a simple counting argument, using the property that $\forall R$, $P_1 = F_1(W(Q), R)$ has at most $|\mathcal{W}|$ choices of values.

When $i \geq 2$, we can merge data points $Q_1, Q_2$ to be a new data point $Q'$ with domain $\mathcal{Q}' = \mathcal{Q}_1 \times \mathcal{Q}_2$. We can construct $P' = (P_1, P_2)$ from function $F_1, F_2$ by sampling $R_1$ to generate $P_1 = F_1(W(Q), R)$, and then sample independent $R_2$ to generate $P_2 = F_2(W(Q), Q_1, R)$. We know that $\Pr_{R_1, R_2}[P' = Q' \mid Q] = \Pr_{R_1}[P_1 = Q_1 \mid Q] \Pr_{R_2}[P_2 = Q_2 \mid Q] = p_1(Q) \cdot p_2(Q)$. The lemma now follows using the following identity:

$$\log(p_1(Q)|\mathcal{Q}_1|) + \log(p_2(Q)|\mathcal{Q}_2|) = \log(p_1(Q)p_2(Q)|\mathcal{Q}_1||\mathcal{Q}_2|) \ . \qquad \square$$

## F.1  Warmup Examples

Let us first see two warmup applications of Lemma F.1.

**Value-only.** Let $g_1, \ldots, g_N \in [T]$, where each $g_i$ is i.i.d. uniformly chosen at random from $[T]$. Think of these as *values*. Suppose a model, parameterized by $W$, is trained on the training data $\mathcal{Z} = (g_1, \ldots, g_N)$. Assume this model, for a given index $i \in [N]$, can generate a random prediction $f_i$ corresponding to $g_i$. We can represent this model as $f_i(W(\mathcal{Z}), R)$, where $R$ denotes the randomness. The cross-entropy loss for this scenario (averaged over all possible training data) is expressed as

$$\mathbf{loss} \stackrel{\text{def}}{=} \mathbb{E}_g[\mathbf{loss}(g)] \stackrel{\text{def}}{=} \mathbb{E}_g \left[ \frac{1}{N} \sum_{i \in [N]} -\log \Pr_{f_i}[f_i = g_i] \right] \geq 0$$

Now we apply Lemma F.1 by setting $\mathcal{Q}_1 = \ldots \mathcal{Q}_N = [T]$, $Q_i = g_i$, and $P_i = f_i$. We have

$$\log |\mathcal{W}| \geq \mathbb{E}_g \left[ \sum_{i \in [N]} \log \Pr_{f_i}[f_i = g_i] + \log T \right] = N \log T - N \mathbb{E}_g \mathbf{loss}(g) = \mathbb{E}_g N \log \frac{T}{e^{\mathbf{loss}(g)}} \ .$$

Changing the base immediately yields a bit complexity lower bound of $\log_2 |\mathcal{W}| \geq N \log_2 \frac{T}{e^{\mathbf{loss}}}$. As the loss approaches zero, this matches the bit complexity upper bound.

**Name-only.** Let $g_1, \ldots, g_N \in [N_0]$ be $N$ distinct elements from $[N_0]$, sampled uniformly at random without replacement, and considered as *names*. Suppose a model $f$, parameterized by $W$, is trained on the dataset $\mathcal{Z} = (g_1, \ldots, g_N)$ to predict a name. We denote this as $f(W(\mathcal{Z}), R)$, where $R$ represents randomness. The cross-entropy loss for this scenario is defined as

$$\mathbf{loss} \stackrel{\text{def}}{=} \mathbb{E}_g[\mathbf{loss}(g)] \stackrel{\text{def}}{=} \mathbb{E}_g \Big[ \frac{1}{N} \sum_{i \in [N]} - \log \Pr_f[f = g_i] \Big] \geq 0$$

To apply Lemma F.1, we define $\mathcal{Q}_1 = [N_0]$, $\mathcal{Q}_2 = [N_0 - 1]$, and continue until $\mathcal{Q}_N = [N_0 - N + 1]$. After uniformly randomly generating $Q_1, \ldots, Q_N$ from $\mathcal{Q}_1, \ldots, \mathcal{Q}_N$, we construct $(g_1, \ldots, g_N) \in [N]^N$ as follows: set $g_1 = Q_1$; for $g_2$, set it to $Q_2$ if $Q_2 < Q_1$, otherwise $g_2 = Q_2 + 1$; and in general, define $g_i$ as the $Q_i$-th smallest element in $[N_0] \setminus \{g_1, \ldots, g_{i-1}\}$. This method provides an alternative way to generate $\mathcal{Z} = (g_1, \ldots, g_N)$, denoted as $\mathcal{Z}(Q)$. For each $i \in [N]$, we define $P_i$ as follows: first, generate $f = f(\mathcal{W}(\mathcal{Z}), R_i)$ using fresh randomness $R_i$. Set $P_i \stackrel{\text{def}}{=} s$ if $f$ is the $s$-th smallest element in $[N_0] \setminus \{g_1, \ldots, g_{i-1}\}$, or a special symbol such as $\varnothing$ if $f$ is among $\{g_1, \ldots, g_{i-1}\}$. (**Note importantly**, this definition of $P_i$ necessitates knowledge of $g_1, \ldots, g_{i-1}$; however, this is permissible as Lemma F.1 allows $P_i$ to depend on $Q_1, \ldots, Q_{i-1}$.) For every fixed $Q$ (and thus fixed $g$),

$$\sum_{i \in [N]} \log \big(p_i(Q)\big) = \sum_{i \in [N]} \log \Big(\Pr_{P_i}[P_i = Q_i]\Big) = \sum_{i \in [N]} \log \Big(\Pr_{R_i}[f(\mathcal{W}(\mathcal{Z}), R_i) = g_i]\Big) = -N\mathbf{loss}(g)$$

Applying Lemma F.1 we have

$$\log |\mathcal{W}| \geq \mathbb{E}_Q \Big[ \sum_{i \in [N]} \log \big(p_i(Q) \times |N_0 - i + 1|\big) \Big] \geq \mathbb{E}_Q \Big[ N \log \frac{N_0 - N}{e^{\mathbf{loss}(g)}} \Big] = N \log \frac{N_0 - N}{e^{\mathbb{E}_Q \mathbf{loss}(g)}} \ .$$

Ideally, if the model $f$ can perfectly memorize the entire training set $\{Z\} = (g_1, \ldots, g_N)$, its best possible loss $\mathbf{loss}(g) = \log N$ is achieved. Thus, if the model can perfectly learn this training set, the bit complexity lower bound satisfies $\log |\mathcal{W}| \geq N \log \frac{N_0 - N}{N} \geq (1 - o(1))N \log \frac{N_0}{N}$ when $N \ll N_0$.

### F.2 Main Proof

We recommend that readers first review the warmup examples in Section F.1 before proceeding with this proof.

*Proof of Theorem 3.2.* Let us first construct the domains $\mathcal{Q}_i$'s in Lemma F.1.

1. Let $\mathcal{Q}_1 = [N_0]$, $\mathcal{Q}_2 = [N_0 - 1] \cdots \mathcal{Q}_N = [N_0 - N + 1]$.
2. Let $\big(\mathcal{Q}_{N+jD+1}, \ldots \mathcal{Q}_{N+jD+D}\big) = \big([T^L], [T^L - 1], \ldots, [T^L - D + 1]\big)$ for every $j = 0, \ldots, K - 1$.
3. Let $\mathcal{Q}_{N+KD+1} = \cdots = \mathcal{Q}_{N+KD+NK} = [D^C]$.

Recall that each $Q_i$ is independently and uniformly generated at random from $\mathcal{Q}_i$. We now present an alternative method for generating the training dataset $\mathcal{Z}(Q)$.

1. Construct $\mathcal{N} = (n_1, \ldots, n_N)$ as follows: Let $n_1$ be the $Q_1$-th name from $\mathcal{N}_0$; for $i > 1$, let $n_i$ be the $Q_i$-th name from $\mathcal{N}_0 \setminus \{n_1, \ldots, n_{i-1}\}$.
2. For each $a' \in [K]$, let $a$ be the $a'$-th attribute in $\mathcal{A}$. Construct $\mathcal{D}_a = (w_1, \ldots, w_D)$ as follows: Let $w_1$ be the $Q_{N+(a'-1)D+1}$-th element in $\mathcal{T}^L$; for $i > 1$, let $w_i$ be the $Q_{N+(a'-1)D+i}$-th element in $\mathcal{T}^L \setminus \{w_1, \ldots, w_{i-1}\}$.

3. For the $n'$-th name $n$ and the $a'$-th attribute $a$, assign its value $v^\star(n,a) = (v_1, \ldots, v_C) \in (\mathcal{D}_a)^C$ by setting each $v_i$ as the $s_i$-th element in $\mathcal{D}_a$, where the integer sequence $(s_1, \ldots, s_C) := Q_{N+KD+(n'-1)K+a'} \in [D^C]$.

It is easy to verify that this gives the same dataset distribution as Definition 2.2. Next, consider $Q$ being fixed (thus the dataset $\mathcal{Z}$ being fixed), we construct $P_1, P_2, \cdots, P_{N+KD+NK}$ using the given model functions $F^\top(W(\mathcal{Z}), R)$ and $F^\perp(W(\mathcal{Z}), n, a, R)$.

**Name part.** For the name part, construct $P_i$ for $i \in [N]$ following the approach from the "value-only" warmup example. Specifically, let $R_i$ be fresh randomness, and define $P_i = s$ if $F^\top(W(\{Z\}), R_i)$ matches the $s$-th element in $\mathcal{N}_0 \setminus \{n_1, \ldots, n_{i-1}\}$, or an arbitrary symbol $\varnothing$ if it falls within $\{n_1, \ldots, n_{i-1}\}$.[29] Adopting the analysis from the "name-only" warmup example, we obtain

$$\sum_{i \in [N]} \log \mathbf{Pr}_{P_i}[P_i = Q_i] = -N \mathbf{loss}_{name}(\mathcal{Z}) \ . \tag{F.2}$$

**Diversity Part.** For the diversity component, we construct the $P_i$'s as follows. For each $a' \in [K]$, let $a$ denote the $a'$-th attribute in $\mathcal{A}$. We form $P_{N+(a'-1)D+i}$ by initially calculating $F^\perp(W(\mathcal{Z}), n, a, R_i)$, where $n \in \mathcal{N}$ is selected uniformly at random. [30] Subsequently, if $F^\perp(W(\mathcal{Z}), n, a, R_i)$ corresponds to the $s$-th element in $\mathcal{T}^L \setminus \{w_1, \ldots, w_{i-1}\}$, then set $P_{N+(a'-1)D+i} = s$; otherwise, set $P_{N+(a'-1)D+i} = \varnothing$.[31]

Now, let $a$ be the $a'$-th element in $\mathcal{A}$. Consider $Q$ as fixed, with randomness arising solely from the calculation of $P_i$'s. Note that $Q$ establishes an order of elements in $\mathcal{D}_a$, denoted by $w_1, \ldots, w_D$. We have

$$\sum_{i \in [D]} \log \mathbf{Pr}_{P_{N+(a'-1)D+i}}[P_{N+(a'-1)D+i} = Q_{N+(a'-1)D+i}] = \sum_{i \in [D]} \log \mathbb{E}_{n \in \mathcal{N}} \mathbf{Pr}_R \left[ F_1^\perp(W(\mathcal{Z}), n, a, R) = w_i \right]$$

$$= \sum_{w \in \mathcal{D}_a} \log \mathbb{E}_{n \in \mathcal{N}} \mathbf{Pr}_R \left[ F_1^\perp(W(\mathcal{Z}), n, a, R) = w \right] =: \spadesuit_a$$

Let us denote by $\mathcal{N}_{w,a}$ the set of $n \in \mathcal{N}$ so that $v^\star(n,a) = w$. We have

$$\spadesuit_a = \sum_{w \in \mathcal{D}_a} \log \sum_{n \in \mathcal{N}} \mathbf{Pr}_R \left[ F_1^\perp(W(\mathcal{Z}), n, a, R) = w \right] - D \log N$$

$$\overset{①}{\geq} \sum_{w \in \mathcal{D}_a} \log \sum_{n \in \mathcal{N}_{w,a}} \mathbf{Pr}_R \left[ F_1^\perp(W(\mathcal{Z}), n, a, R) = w \right] - D \log N$$

$$= \sum_{w \in \mathcal{D}_a} \log \frac{1}{|\mathcal{N}_{w,a}|} \sum_{n \in \mathcal{N}_{w,a}} \mathbf{Pr}_R \left[ F_1^\perp(W(\mathcal{Z}), n, a, R) = w \right] - D \log N + \sum_{w \in \mathcal{D}} \log |\mathcal{N}_{w,a}|$$

$$\overset{②}{\geq} \sum_{w \in \mathcal{D}_a} \frac{1}{|\mathcal{N}_{w,a}|} \sum_{n \in \mathcal{N}_{w,a}} \log \mathbf{Pr}_R \left[ F_1^\perp(W(\mathcal{Z}), n, a, R) = w \right] - D \log N + \sum_{w \in \mathcal{D}} \log |\mathcal{N}_{w,a}|$$

Above, ① uses monotonicity of the log function and ② uses convexity of the log function. Using simple Chernoff bound, one can see that as long a $N \geq \Omega(D \log N)$, with high probability $|\mathcal{N}_{w,a}| \geq$

---

[29]Importantly, $P_i$ may depend on $n_1, \ldots, n_{i-1}$; however, since Lemma F.1 permits $P_i$ to depend on $Q_1, \ldots, Q_{i-1}$, this is acceptable.

[30]Importantly, $P_{N+(a'-1)D+i}$ depends on $\mathcal{N}$; however, since Lemma F.1 permits $P_i$ to depend on $Q_1, \ldots, Q_{i-1}$, and since $\mathcal{N}$ is uniquely determined by $Q_1, \ldots, Q_N$, this is acceptable.

[31]Importantly, $P_{N+(a'-1)D+i}$ depends on $w_1, \ldots, w_{i-1}$; however, since Lemma F.1 permits $P_i$ to depend on $Q_{N+(a'-1)D+1}, \ldots, Q_{N+(a'-1)D+i-1}$, this is acceptable.

$(1 - o(1))\frac{N}{D}$ for all $w \in D$. Thus, we know with high probability

$$\spadesuit_a \geq (1 + o(1))D \sum_{w \in \mathcal{D}_a} \frac{1}{N} \sum_{n \in \mathcal{N}_{w,a}} \log \Pr_R \left[ F_1^\perp(W(\mathcal{Z}), n, a, R) = w \right] - D \log D - o(D)$$

$$= (1 + o(1))D \frac{1}{N} \sum_{n \in \mathcal{N}} \log \Pr_R \left[ F_1^\perp(W(\mathcal{Z}), n, a, R) = v^\star(n, a) \right] - D \log D - o(D)$$

Thus, summing up over all the diversity part, we have (recall we are fixing $Q$ and thus fixing $\mathcal{Z}$)

$$\sum_{i \in [KD]} \log \Pr_{P_{N+i}} [P_{N+i} = Q_{N+i}]$$

$$\geq (1 + o(1))D \frac{1}{NK} \sum_{n \in \mathcal{N}, a \in \mathcal{A}} \log \Pr_R \left[ F_1^\perp(W(\mathcal{Z}), n, a, R) = v^\star(n, a) \right] - KD \log D - o(KD)$$

$$= -(1 + o(1))D \mathbf{loss}_{value1}(\mathcal{Z}) - KD \log D - o(KD) \ . \tag{F.3}$$

**Value part.** For the value part, we construct $P_{N+KD+1}, \ldots, P_{N+KD+NK}$ as follows. For $P_{N+KD+(n'-1)K+a'}$, letting $n$ be the $n'$-th name in $\mathcal{N}$ and $a$ be the $a'$-th attribute in $\mathcal{A}$. Let us compute $F^\perp(W(\mathcal{Z}), n, a, R)$ and find the corresponding $s_1, \ldots, s_C \in [D]$ such that $F_i^\perp(W(\mathcal{Z}), n, a, R)$ is the $s_i$-th element in $\mathcal{D}_a$ for each $i \in [C]$. If not found, we define $P_{N+KD+(n'-1)K+a'} = \varnothing$; otherwise, define $P_{N+KD+(n'-1)K+a'} = (s_1, \ldots, s_C) \in [D^C].$[32]

Following the same simple argument as the "value-only" warmup example, we have

$$\sum_{i \in [NK]} \log \Pr_{P_{N+KD+i}} [P_{N+KD+i} = Q_{N+KD+i}] = \sum_{n \in \mathcal{N}, a \in \mathcal{A}} \Pr \left[ F^\perp(W(\mathcal{Z}), n, a, R) = v^\star(n, a) \right]$$

$$= -NK \mathbf{loss}_{value}(\mathcal{Z}) \quad \text{(F.4)}$$

Summing (F.2) (F.3) and (F.4), and applying Lemma F.1, we have

$$\log |\mathcal{W}| \geq \mathbb{E}_{\mathcal{Z}} \left[ N \log \frac{N_0 - N}{e^{\mathbf{loss}_{name}(\mathcal{Z})}} + NK \log \frac{D^C}{e^{\mathbf{loss}_{value}(\mathcal{Z})}} + KD \log \frac{T^L - D}{De^{(1+o(1))\mathbf{loss}_{value1}(\mathcal{Z})}} - o(KD) \right] \ .$$

This finishes the proof of Theorem 3.2. □

## G  Missing Remark

*Remark* G.1. Due to the significant overlap among textbooks, especially those designed for PreK-12 education, estimating the total amount of knowledge contained within all English-language textbooks can be challenging. However, we attempt to do so as follows.

According to a 2023 article, Pearson Education, a UK-based educational publisher, reported the highest revenue in 2021, with Wiley and McGraw Hill being the top two US-based educational publishers in terms of revenue.[33]

- Pearson's official website lists fewer than 2,100 textbooks.[34]

---

[32]Again, importantly, we can do so because $P_{N+KD+(n'-1)K+a'}$ depends on $\mathcal{N}, \mathcal{D}_a$ but they can be computed using the values of $Q_1, \ldots, Q_{N+KD}$.

[33]https://wordsrated.com/education-book-publishing-companies-statistics/, accessed March 2024.

[34]https://www.pearson.com/en-us/pearsonplus/search.html for their full list of eTextbooks and http://www.mypearsonstore.com/bookstore/browse.asp for their full list of hard copy books, both accessed March 2024.

- Wiley's official website lists fewer than 69,000 textbooks.[35]
- McGraw Hill lists fewer than 22,000 textbooks for PreK-12 education, many of which have significant content overlap (as many are tailored for one of the 50 US states).[36] They list fewer than 2,000 textbooks for higher education.

Taking these figures into account, it seems reasonable to estimate that the content of all English-language textbooks could be condensed into no more than 100,000 textbooks. Assuming an average of 160,000 words per book (e.g., 400 pages with 400 words each), this would amount to a total of 16 billion words.

# References

[1] Ibrahim M Alabdulmohsin, Behnam Neyshabur, and Xiaohua Zhai. Revisiting neural scaling laws in language and vision. *Advances in Neural Information Processing Systems*, 35:22300–22312, 2022.

[2] Zeyuan Allen-Zhu and Yuanzhi Li. Physics of Language Models: Part 1, Context-Free Grammar. *ArXiv e-prints*, abs/2305.13673, May 2023. Full version available at http://arxiv.org/abs/2305.13673.

[3] Zeyuan Allen-Zhu and Yuanzhi Li. Physics of Language Models: Part 3.1, Knowledge Storage and Extraction. *ArXiv e-prints*, abs/2309.14316, September 2023. Full version available at http://arxiv.org/abs/2309.14316.

[4] Zeyuan Allen-Zhu and Yuanzhi Li. Physics of Language Models: Part 3.2, Knowledge Manipulation. *ArXiv e-prints*, abs/2309.14402, September 2023. Full version available at http://arxiv.org/abs/2309.14402.

[5] Zeyuan Allen-Zhu, Yuanzhi Li, and Yingyu Liang. Learning and generalization in overparameterized neural networks, going beyond two layers. *Advances in neural information processing systems*, 32, 2019.

[6] Zeyuan Allen-Zhu, Yuanzhi Li, and Zhao Song. A convergence theory for deep learning via over-parameterization. In *International conference on machine learning*, pages 242–252. PMLR, 2019.

[7] Sid Black, Stella Biderman, Eric Hallahan, Quentin Anthony, Leo Gao, Laurence Golding, Horace He, Connor Leahy, Kyle McDonell, Jason Phang, Michael Pieler, USVSN Sai Prashanth, Shivanshu Purohit, Laria Reynolds, Jonathan Tow, Ben Wang, and Samuel Weinbach. GPT-NeoX-20B: An open-source autoregressive language model. In *Proceedings of the ACL Workshop on Challenges & Perspectives in Creating Large Language Models*, 2022. URL https://arxiv.org/abs/2204.06745.

[8] Sébastien Bubeck, Varun Chandrasekaran, Ronen Eldan, Johannes Gehrke, Eric Horvitz, Ece Kamar, Peter Lee, Yin Tat Lee, Yuanzhi Li, Scott Lundberg, et al. Sparks of artificial general intelligence: Early experiments with gpt-4. *arXiv preprint arXiv:2303.12712*, 2023.

[9] William Fedus, Barret Zoph, and Noam Shazeer. Switch transformers: Scaling to trillion parameter models with simple and efficient sparsity. *The Journal of Machine Learning Research*, 23(1):5232–5270, 2022.

[10] Elias Frantar, Saleh Ashkboos, Torsten Hoefler, and Dan Alistarh. GPTQ: Accurate post-training compression for generative pretrained transformers. *arXiv preprint arXiv:2210.17323*, 2022.

[11] Olga Golovneva, Zeyuan Allen-Zhu, Jason Weston, and Sainbayar Sukhbaatar. Reverse training to nurse the reversal curse. *arXiv preprint arXiv:2403.13799*, 2024.

[12] Suriya Gunasekar, Yi Zhang, Jyoti Aneja, Caio César Teodoro Mendes, Allie Del Giorno, Sivakanth Gopi, Mojan Javaheripi, Piero Kauffmann, Gustavo de Rosa, Olli Saarikivi, et al. Textbooks are all you need. *arXiv preprint arXiv:2306.11644*, 2023.

---

[35] https://www.wiley.com/en-us/subjects, accessed March 2024. We wrote a code to sum up all the books in all of their subcategories; our code may double count books, so this is only a safe upper bound. We used this number instead of the "21,000" online books mentioned on https://www.wiley.com/learn/librarysolutions/online-books-purchase.html, accessed March 2024.

[36] https://www.mheducation.com/search.html?searchQuery=&page=1&sortby=title_desc&order=desc&bu=seg&TYPE=Products&PRODUCT_TYPE_PATH=_Student+Materials, accessed March 2024.

[13] Tom Henighan, Jared Kaplan, Mor Katz, Mark Chen, Christopher Hesse, Jacob Jackson, Heewoo Jun, Tom B Brown, Prafulla Dhariwal, Scott Gray, et al. Scaling laws for autoregressive generative modeling. *arXiv preprint arXiv:2010.14701*, 2020.

[14] Danny Hernandez, Jared Kaplan, Tom Henighan, and Sam McCandlish. Scaling laws for transfer. *arXiv preprint arXiv:2102.01293*, 2021.

[15] Joel Hestness, Sharan Narang, Newsha Ardalani, Gregory Diamos, Heewoo Jun, Hassan Kianinejad, Md Mostofa Ali Patwary, Yang Yang, and Yanqi Zhou. Deep learning scaling is predictable, empirically. *arXiv preprint arXiv:1712.00409*, 2017.

[16] Jordan Hoffmann, Sebastian Borgeaud, Arthur Mensch, Elena Buchatskaya, Trevor Cai, Eliza Rutherford, Diego de Las Casas, Lisa Anne Hendricks, Johannes Welbl, Aidan Clark, et al. Training compute-optimal large language models. *arXiv preprint arXiv:2203.15556*, 2022.

[17] Edward J Hu, Phillip Wallis, Zeyuan Allen-Zhu, Yuanzhi Li, Shean Wang, Lu Wang, Weizhu Chen, et al. LoRA: Low-Rank Adaptation of Large Language Models. In *ICLR*, 2021.

[18] Changho Hwang, Wei Cui, Yifan Xiong, Ziyue Yang, Ze Liu, Han Hu, Zilong Wang, Rafael Salas, Jithin Jose, Prabhat Ram, Joe Chau, Peng Cheng, Fan Yang, Mao Yang, and Yongqiang Xiong. Tutel: Adaptive mixture-of-experts at scale. *CoRR*, abs/2206.03382, June 2022. URL https://arxiv.org/pdf/2206.03382.pdf.

[19] Albert Q Jiang, Alexandre Sablayrolles, Arthur Mensch, Chris Bamford, Devendra Singh Chaplot, Diego de las Casas, Florian Bressand, Gianna Lengyel, Guillaume Lample, Lucile Saulnier, et al. Mistral 7b. *arXiv preprint arXiv:2310.06825*, 2023.

[20] Mandar Joshi, Eunsol Choi, Daniel S Weld, and Luke Zettlemoyer. Triviaqa: A large scale distantly supervised challenge dataset for reading comprehension. *arXiv preprint arXiv:1705.03551*, 2017.

[21] Jared Kaplan, Sam McCandlish, Tom Henighan, Tom B Brown, Benjamin Chess, Rewon Child, Scott Gray, Alec Radford, Jeffrey Wu, and Dario Amodei. Scaling laws for neural language models. *arXiv preprint arXiv:2001.08361*, 2020.

[22] Tom Kwiatkowski, Jennimaria Palomaki, Olivia Redfield, Michael Collins, Ankur Parikh, Chris Alberti, Danielle Epstein, Illia Polosukhin, Jacob Devlin, Kenton Lee, et al. Natural questions: a benchmark for question answering research. *Transactions of the Association for Computational Linguistics*, 7:453–466, 2019.

[23] Yuanzhi Li and Yingyu Liang. Learning overparameterized neural networks via stochastic gradient descent on structured data. In *Advances in Neural Information Processing Systems*, 2018.

[24] Yuanzhi Li, Sébastien Bubeck, Ronen Eldan, Allie Del Giorno, Suriya Gunasekar, and Yin Tat Lee. Textbooks are all you need ii: phi-1.5 technical report. *arXiv preprint arXiv:2309.05463*, 2023.

[25] Niklas Muennighoff, Alexander M Rush, Boaz Barak, Teven Le Scao, Aleksandra Piktus, Nouamane Tazi, Sampo Pyysalo, Thomas Wolf, and Colin Raffel. Scaling data-constrained language models. *arXiv preprint arXiv:2305.16264*, 2023.

[26] Alec Radford, Jeffrey Wu, Rewon Child, David Luan, Dario Amodei, Ilya Sutskever, et al. Language models are unsupervised multitask learners. *OpenAI blog*, 1(8):9, 2019.

[27] Jonathan S Rosenfeld. Scaling laws for deep learning. *arXiv preprint arXiv:2108.07686*, 2021.

[28] Jonathan S Rosenfeld, Amir Rosenfeld, Yonatan Belinkov, and Nir Shavit. A constructive prediction of the generalization error across scales. *arXiv preprint arXiv:1909.12673*, 2019.

[29] Noam Shazeer. Glu variants improve transformer. *arXiv preprint arXiv:2002.05202*, 2020.

[30] Noam Shazeer, Azalia Mirhoseini, Krzysztof Maziarz, Andy Davis, Quoc Le, Geoffrey Hinton, and Jeff Dean. Outrageously large neural networks: The sparsely-gated mixture-of-experts layer. In *International Conference on Learning Representations*, 2016.

[31] Jianlin Su, Yu Lu, Shengfeng Pan, Bo Wen, and Yunfeng Liu. Roformer: Enhanced transformer with rotary position embedding, 2021.

[32] Hugo Touvron, Thibaut Lavril, Gautier Izacard, Xavier Martinet, Marie-Anne Lachaux, Timothée Lacroix, Baptiste Rozière, Naman Goyal, Eric Hambro, Faisal Azhar, et al. Llama: Open and efficient

foundation language models. *arXiv preprint arXiv:2302.13971*, 2023.

[33] Hugo Touvron, Louis Martin, Kevin Stone, Peter Albert, Amjad Almahairi, Yasmine Babaei, Nikolay Bashlykov, Soumya Batra, Prajjwal Bhargava, Shruti Bhosale, et al. Llama 2: Open foundation and fine-tuned chat models. *arXiv preprint arXiv:2307.09288*, 2023.

[34] Dingli Yu, Simran Kaur, Arushi Gupta, Jonah Brown-Cohen, Anirudh Goyal, and Sanjeev Arora. Skill-mix: A flexible and expandable family of evaluations for ai models. *arXiv preprint arXiv:2310.17567*, 2023.